\documentclass[11pt]{article}

\usepackage[final]{acl}
\usepackage{comment}
\usepackage{float}
\usepackage{listings}

\usepackage{times}
\usepackage{latexsym}

\usepackage[T1]{fontenc}

\usepackage[utf8]{inputenc}

\usepackage{microtype}

\usepackage{inconsolata}

\usepackage{graphicx}
\usepackage{booktabs}
\usepackage{multirow}
\usepackage{array}
\usepackage{tabularx}
\usepackage{makecell}
\usepackage{amsmath,amssymb,amsfonts}
\usepackage[table]{xcolor}
\usepackage{xspace}
\usepackage{url}
\usepackage{enumitem}
\usepackage{placeins}
\usepackage{authblk}
\usepackage{dcolumn}
\usepackage[normalem]{ulem}
\usepackage{placeins}
\def\secref#1{\S\ref{sec:#1}}
\def\seclabel#1{\label{sec:#1}}

\newcommand{\datg}{DATG\xspace}

\newcommand{\teaserfig}{Picture_Table/datg_intro.pdf}
\newcommand{\overviewfig}{Picture_Table/datg_main.pdf}

\definecolor{GoodA}{HTML}{D8F0D6}
\definecolor{GoodB}{HTML}{EAF6DD}
\definecolor{MidC}{HTML}{FFF3C7}
\definecolor{WarnD}{HTML}{FADBC8}
\definecolor{BadE}{HTML}{F4B8B0}

\newcolumntype{Y}{>{\raggedright\arraybackslash}X}
\newcolumntype{L}[1]{>{\raggedright\arraybackslash}p{#1}}

\title{Beyond Input Understanding: Diagnosing Multilingual Mathematical Reasoning with Directed Acyclic Trace Graphs}

\author{
 \textbf{Jiaqiao Zhang\textsuperscript{1}},
 \textbf{Zhuojun Li\textsuperscript{1}},
 \textbf{Raoyuan Zhao\textsuperscript{2,3}},\\
 \textbf{Jian Lan\textsuperscript{2,3}},
 \textbf{Thomas Seidl\textsuperscript{2,3}},
 \textbf{Michael A. Hedderich\textsuperscript{2,3}},
 \textbf{Hinrich Sch\"utze\textsuperscript{2,3}},
 \textbf{Yihong Liu\textsuperscript{2,3,\dag}}
\\
\\
 \textsuperscript{1}Southwest University \space
 \textsuperscript{2}LMU Munich \\
 \textsuperscript{3}Munich Center for Machine Learning (MCML)
\\
 \small{
 \texttt{wh2149972901@email.swu.edu.cn}
 \space \space \space
 \texttt{\{rzhao, hedderich, yihong\}@cis.lmu.de}
 }
}

\begin{document}
\sloppy
\maketitle

\def\thefootnote{$\dag$}\footnotetext{Corresponding author.}\def\thefootnote{\arabic{footnote}}

\begin{abstract}
Large reasoning models (LRMs) achieve strong mathematical reasoning performance in English, but remain much less reliable in many low- and medium-resource languages. 
This gap is often explained as a failure to \emph{understand} non-English problem statements. 
We show that this view is incomplete: even when the problem is given in English, controlling the model's reasoning language can substantially reduce accuracy, suggesting that language also affects \emph{reasoning execution} itself.
To study this effect, we introduce \textbf{\datg}, a \textbf{D}irected \textbf{A}cyclic \textbf{T}race \textbf{G}raph framework that maps reasoning traces to language-independent mathematical \emph{anchors} and \emph{dependencies}. 
This allows us to align target-language traces with reference DAGs and measure whether they cover required mathematical nodes, respect dependency edges, and avoid harmful mathematical actions.
Experiments on the Qwen3 series across 12 languages show that non-English reasoning often suffers from reduced anchor coverage and weaker dependency fidelity, especially in low-resource languages.
Motivated by this diagnosis, we propose Loop-Retry and Formula-Retry, two simple test-time controls targeting DATG-exposed failure modes, and show that they consistently improve target-language reasoning performance in low-resource languages.\footnote{Our code and data are available at \url{https://github.com/Pluto-R/DATG}.}
\end{abstract}

\section{Introduction}

Chain-of-thought (CoT) reasoning has greatly improved the ability of large reasoning models (LRMs) to solve multi-step mathematical problems \citep{wei2022cot}. 
However, these gains are not evenly distributed across languages. 
Models that perform well on English problems often suffer large performance drops when the same problems are presented in medium- or low-resource languages such as Bengali, Swahili, or Telugu \citep{shi2023mgsm,lai2024mcot,wang2025polymath,luo2025mmath,liu2026largereasoning}. 

\begin{figure}[t]
  \centering
  \setlength{\belowcaptionskip}{0.2cm}
\includegraphics[width=0.95\columnwidth]{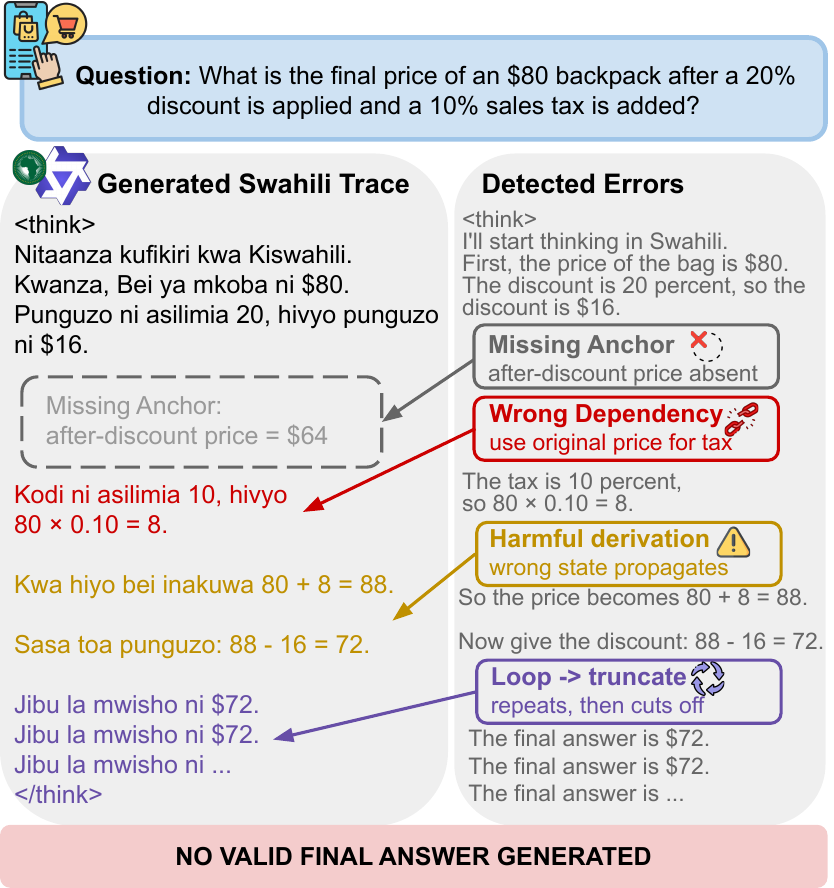}
  \caption{Example of reasoning execution failure under the $en\rightarrow sw$ setting. 
  Although the problem is given in English, the Swahili reasoning trace misses the after-discount price anchor, computes tax from the original price, propagates the resulting error, and eventually falls into a loop without producing a valid final answer.}
  \label{fig:teaser}
\end{figure}

A common explanation is that such gaps mainly come from poor \emph{understanding} of non-English problem statements \citep{etxaniz-etal-2024-multilingual,liu-etal-2025-translation,kang2026gaps}. 
That is, if a model misreads the problem, the reasoning process is likely to fail before it begins.
While this explanation is important, we argue that it is incomplete. 
Even when the problem is kept in English, requesting a reasoning trace
in another language can produce a substantial performance drop, especially in
low- or medium-resource languages
\citep{qi2025control,wang2025mixing,tam2025language,barua2026long,yong2025crosslingual,zhao2026comprehensive,liu2026largereasoning}.
This suggests that multilingual reasoning gaps are not only input-side understanding failures, but also trace-side \emph{reasoning execution} failures: the requested reasoning language may affect how LRMs derive intermediate facts, maintain dependencies, and reach a final answer. 
As illustrated in Figure~\ref{fig:teaser}, such failures can appear even when the problem is given in English. 
Despite this, relatively little work has examined how reasoning execution differs across languages at the trace level.\footnote{A practical alternative is to reason in English and then produce a target-language response. While effective for answer delivery, this addresses a different setting from the one studied here: we focus on whether the reasoning process itself remains reliable when conducted directly in the target language. Users may need to inspect and understand intermediate reasoning steps, and we cannot assume English proficiency. This is particularly relevant in settings such as education and auditing, where reasoning is itself part of the user-facing output.}

We propose the \textbf{D}irected \textbf{A}cyclic \textbf{T}race \textbf{G}raph framework (\textbf{\datg}), a diagnostic framework that abstracts reasoning traces into language-independent mathematical anchors and dependency edges.
Anchors represent checkable reasoning units such as equations, derived quantities, relations, or intermediate results; edges encode prerequisite structure.
By aligning a target-language reasoning trace to these anchors and dependencies, \datg allows us to examine whether the target-language trace commits the required mathematical anchors, preserves the extracted dependency structure, and avoids harmful mathematical actions.

Using \datg, we study three Qwen3 models across 12 languages.
Our analysis shows that non-English reasoning often degrades both anchor coverage and dependency fidelity, with the strongest failures appearing in medium- and low-resource languages. 
These failures indicate that low-resource target-language reasoning traces often lose part of the derivational structure needed for correct mathematical reasoning: they may omit key intermediate facts, break dependency order, or introduce harmful mathematical actions.

Motivated by these diagnostic findings, we further ask whether DATG-exposed failure modes can be mitigated with simple test-time controls. 
We propose \emph{Loop-Retry}, which resamples when early repetitive degeneration is detected, and \emph{Formula-Retry}, which provides a compact symbolic scaffold of known quantities, relations, and formula templates without revealing the final result. 
Both controls improve low-resource reasoning performance.

Our contributions are as follows: 
(\textbf{i}) We separate input-side understanding from trace-side reasoning execution and show that reasoning language alone can cause large performance drops (\secref{accuracy}).
(\textbf{ii}) We introduce \datg, a reference-conditioned graph diagnostic that maps
reasoning traces to language-independent mathematical anchors and dependencies (\secref{datg}).
(\textbf{iii}) We 
conduct a DATG-based analysis across 12 languages on the Qwen3 series, showing that low-resource languages typically lose anchors, break dependency order, and introduce more harmful mathematical actions, especially on 
harder problems
(\secref{findings}).
(\textbf{iv}) Motivated by the \datg diagnosis, we propose Loop-Retry and Formula-Retry as lightweight test-time controls, showing that DATG-exposed failure modes can be partially mitigated in low-resource target-language traces (\secref{control}).

\section{Related Work}
\paragraph{Multilingual Reasoning.}

Chain-of-thought prompting improves mathematical reasoning, but its gains are uneven across languages, difficulty levels, and reasoning formats \citep{shi2023mgsm,lai2024mcot,wang2025polymath,luo2025mmath,zhao2026comprehensive,ghosh2025survey}. 
Recent work further shows that LRMs are sensitive not only to the input language, but also to the reasoning-trace language \citep{qi2025control,wang2025mixing,tam2025language,barua2026long,yong2025crosslingual,liu2026largereasoning}. 
A common explanation focuses on input-side understanding: models may fail because they misread non-English problem statements \citep{kang2026gaps}. 
Our work keeps this explanation in view, but separates it from trace-side \emph{reasoning execution}: we hold the input problem fixed in English and examine how target-language reasoning traces structurally differ.

Prior work has improved multilingual reasoning through several routes:
cross-lingual prompting
\citep{qin2023crosslingual,huang2023notall};
question alignment and 
instruction tuning \citep{zhu2024question,lai2024mcot};
augmentation by external multilingual understanding modules \citep{huang2024mindmerger};
preference optimization or reinforcement learning \citep{she2024mapo,zhang2025thinknatively};
self-training
\citep{ranaldi2025multilingual};
and crosslingual on-policy self-distillation from high-resource reasoning behavior \citep{liu2026copsd}.

These methods are complementary to ours. 
Rather than training a stronger multilingual model, we diagnose how reasoning execution breaks at the trace level by comparing language-independent mathematical anchors and dependencies.

\paragraph{Process-Level Diagnosis of Reasoning.}

Final-answer accuracy is often too coarse to localize mathematical reasoning failures, since the same outcome may hide missing intermediate steps, invalid transformations, broken dependencies, or accidentally correct answers. 
Prior work therefore studies richer process signals: solution verifiers rank complete candidate solutions \citep{cobbe2021training}, process supervision provides step-level feedback \citep{lightman2024verify}, reasoning-trace metrics evaluate coherence and step validity \citep{golovneva2023roscoe}, and step-error benchmarks test whether models can identify where a solution goes wrong \citep{zheng2025processbench}. 
Recent work further surveys trace-evaluation criteria such as factuality, validity, coherence, and utility \citep{lee-hockenmaier-2025-evaluating}, while structural and graph-based methods organize reasoning into intermediate units and dependencies \citep{besta2024graph,lee2025reasoningflowsemanticstructurecomplex}. 
Interpretability work also suggests that reasoning traces contain meaningful internal structure, including causally influential reasoning steps and thought anchors \citep{paul-etal-2024-making,bogdan2025thoughtanchors}. 
In multilingual settings, related work studies process reward modeling for multilingual multi-step reasoning and compares English and multilingual traces using features such as language alignment, step counts, and reasoning flow \citep{wang2025demystifying,ki2026what}.

Our work builds on this process-level view but focuses on cross-lingual differences in trace-side \emph{reasoning execution}. 
\datg aligns target-language reasoning traces to English reference DAGs with language-independent mathematical anchors and dependency edges.
This enables us to audit whether multilingual traces cover required anchors, preserve prerequisite relations, and introduce harmful mathematical actions.

\section{Experimental Setup}\seclabel{setup}

\begin{figure*}[t]
  \centering
\setlength{\belowcaptionskip}{0.4cm}
  \includegraphics[
    width=0.98\textwidth,
    height=0.36\textheight,
  ]{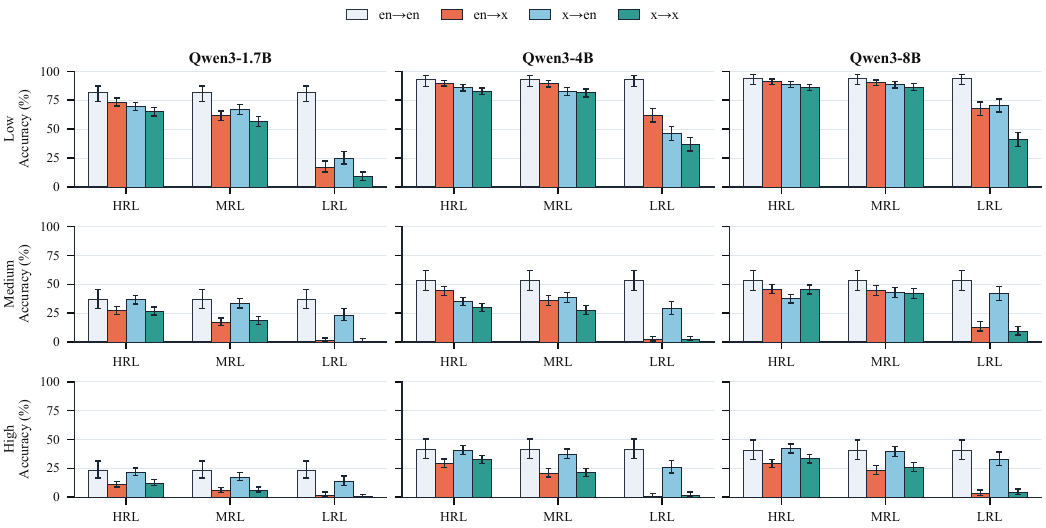}
        \caption{
        Final-answer accuracy across input--reasoning language settings. 
        Each panel shows one model--difficulty pair; foreground bars report averaged accuracy over target languages in each resource group, with 95\% Wilson confidence intervals. 
        The pale bar
        i.e., $en\rightarrow en$ 
        is not included in resource-group averages. 
        The drop from $en\rightarrow en$ to $en\rightarrow x$ shows that changing only the reasoning-trace language can substantially reduce accuracy.
        }
  \label{fig:execution_gap}
\end{figure*}

\subsection{Input-Reasoning Language Settings}
\seclabel{language-settings}

We evaluate four input-reasoning language settings.
The $en\rightarrow en$ setting uses English for both the problem and the reasoning trace, and serves as the reference condition.
The $en\rightarrow x$ setting keeps the problem in English but asks the model to generate the reasoning trace in target language $x$.
The $x\rightarrow en$ setting changes the input language while keeping the reasoning trace in English.
The $x\rightarrow x$ setting uses the target language for both input and reasoning.
Together, these settings help disentangle input-side understanding from trace-side reasoning execution. 
Throughout the paper, we use \emph{reasoning execution} to refer to the model's visible, generated reasoning trace under a specified language.

\subsection{Models, Languages, and Benchmark}

Our main experiments evaluate \texttt{Qwen3-1.7B}, \texttt{Qwen3-4B}, and \texttt{Qwen3-8B} \citep{yang2025qwen3}. 
To test whether the observed trace-side degradation can be generalized beyond the Qwen3 model family, we additionally evaluate \texttt{DeepSeek-R1-Distill-Llama-8B} and \texttt{DeepSeek-R1-Distill-Qwen-32B}. 
Our language set includes English (\textbf{en}), French (\textbf{fr}), Russian (\textbf{ru}), Chinese (\textbf{zh}), Japanese (\textbf{ja}), Thai (\textbf{th}), Korean (\textbf{ko}), Indonesian (\textbf{id}), Malay (\textbf{ms}), Swahili (\textbf{sw}), Bengali (\textbf{bn}), and Telugu (\textbf{te}).
We treat English, French, Chinese, Japanese, Russian, and Korean as high-resource languages (HRLs); 
Indonesian, Malay, Thai, and Bengali as medium-resource languages (MRLs); 
and Swahili and Telugu as low-resource languages (LRLs), reflecting differences in web-scale text availability, multilingual NLP resources, and model training coverage \citep{joshi-etal-2020-state,nigatu-etal-2024-zenos}.

We use \textbf{PolyMath}, a challenging multilingual mathematical reasoning benchmark with problems spanning multiple difficulty levels \citep{wang2025polymath}. 
For each difficulty level, PolyMath provides 125 questions per language. 
In this study, we evaluate the \textbf{low}, \textbf{medium}, and \textbf{high} difficulty subsets and report results separately for each.

\subsection{Language Control}

The main experiments use a target-language prefix.\!\footnote{Full prompt templates and language compliance statistics for the main $en\rightarrow x$ setting are provided in Appendix~\ref{app:prompts}.} 
Prior work shows that reasoning language can be effectively guided by prompting
\citep{qi2025control,tam2025language,zhao2026comprehensive,barua2026long,liu2026largereasoning}.
Therefore, we use a language-specific prefix to control the reasoning language,
e.g., for Swahili, we use ``\texttt{<think>\textbackslash n Nitaanza kufikiri kwa Kiswahili.\textbackslash n Kwanza,}'', which means ``\emph{I will start thinking in Swahili. First,}''.

\subsection{Evaluation Metrics}

For the input--reasoning language analysis and the main \datg diagnosis (\secref{accuracy} and \secref{findings}), we report final-answer accuracy by comparing the model's generation with the official PolyMath answer using \texttt{Math-Verify} \citep{mathverify2025}.
We also evaluate reasoning trace quality with the three \datg metrics: CAR, PMF, and HAR (defined in \secref{datg-metrics}).




\begin{table}[t]
\setlength{\belowcaptionskip}{0.3cm}
\centering
\scriptsize
\setlength{\tabcolsep}{2.4pt}
\renewcommand{\arraystretch}{1.08}
\resizebox{\columnwidth}{!}{%
\begin{tabular}{lrrrr}
\toprule
Model & en-en & HRL & MRL & LRL \\
\midrule
\texttt{Qwen3-8B} & 53.6 & 45.9 & 44.8 & 13.2 \\
\texttt{DeepSeek-R1-Distill-Llama-8B} & 33.6 & 29.9 & 17.8 & 10.4 \\
\texttt{DeepSeek-R1-Distill-Qwen-32B} & 50.4 & 46.4 & 39.0 & 12.8 \\
\bottomrule
\end{tabular}}
\caption{Medium-difficulty accuracy on PolyMath. Values are percentages; HRL, MRL, and LRL denote target-language resource groups under $en\rightarrow x$.}
\label{tab:generalization_acc}
\end{table}

\section{Accuracy Gaps Across Input-Reasoning Language Settings}
\seclabel{accuracy}

\begin{figure*}[t]
  \centering
    \setlength{\belowcaptionskip}{0.3cm}
  \includegraphics[
    width=0.95\textwidth
  ]{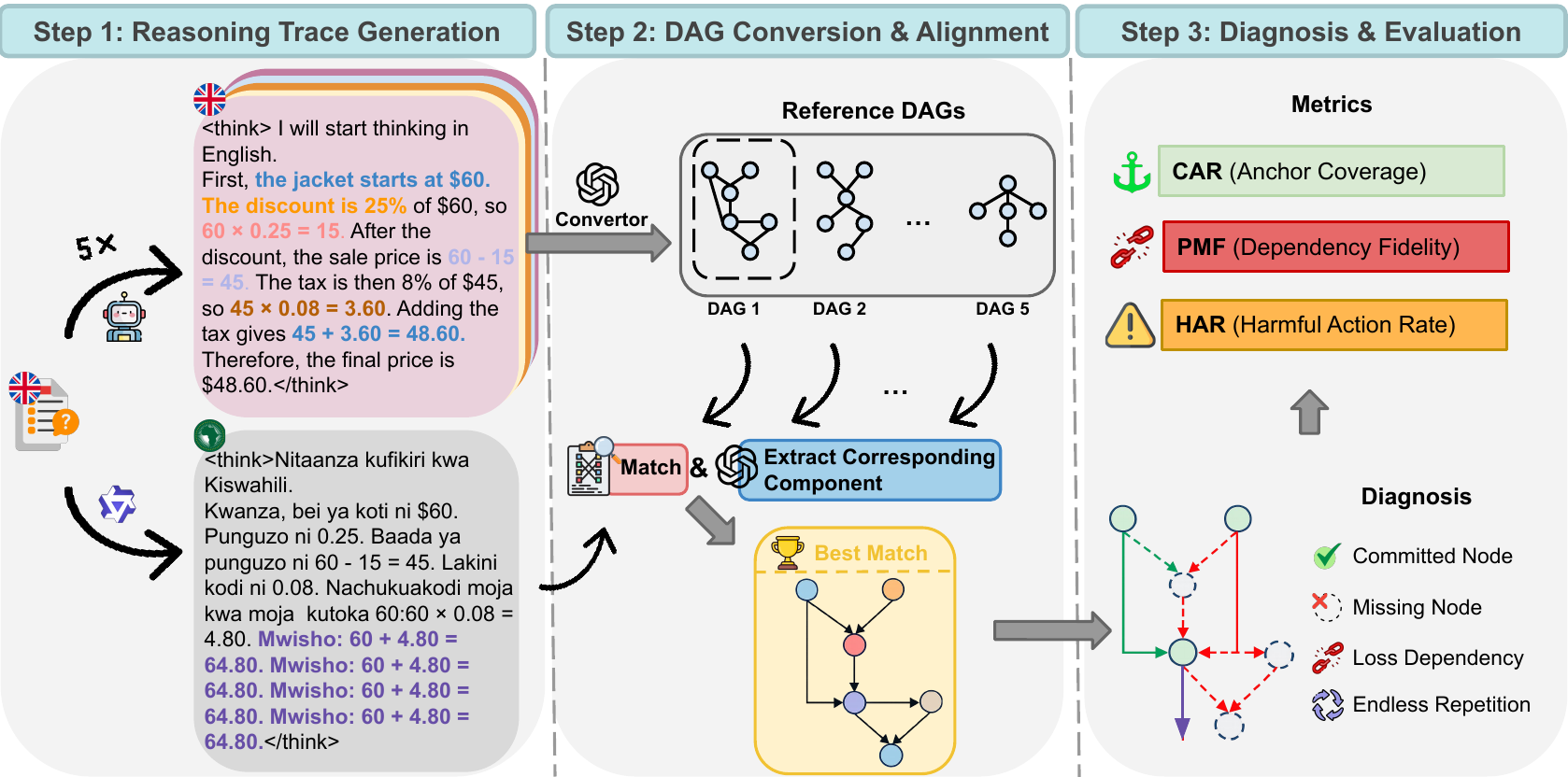}
\caption{
Overview of the \datg diagnosis framework: English reference solutions are converted into reference DAGs, target-language reasoning traces are aligned to English reference DAGs, and the best-aligned structure is used to compute CAR, PMF, and HAR for diagnosing anchor coverage, dependency fidelity, and harmful actions.
}
  \label{fig:overview}
\end{figure*}

Figure~\ref{fig:execution_gap} compares the four-direction analysis settings defined in \secref{setup}, separating the effects of the input language and the reasoning language; Appendix~\ref{app:direction-accuracy} reports the full per-language results.

\textbf{Reasoning language independently affects accuracy.}
The English reference remains the strongest setting in most comparisons.
However, the gap between $en\rightarrow en$ and $en\rightarrow x$ depends strongly on the target-language resource group.
For \texttt{Qwen3-4B} on medium-difficulty problems, accuracy is 53.6\% in $en\rightarrow en$.
The high-resource $en\rightarrow x$ average remains 44.5\%, but the low-resource $en\rightarrow x$ average drops to 2.0\%.
In the same low-resource group, $x\rightarrow en$ reaches 29.2\%.
This contrast makes the trace-side effect explicit: performance is strongly influenced by the language of the reasoning trace, not only by the input language.
The same pattern appears for \texttt{Qwen3-8B}: on high-difficulty low-resource problems, $x\rightarrow en$ reaches 33.2\%, while $en\rightarrow x$ remains 3.2\%.

\textbf{Low-resource reasoning degrades sharply on harder problems.}
Accuracy decreases with problem difficulty in all settings, but the decline is much steeper when the reasoning trace is in a low-resource language.
For \texttt{Qwen3-4B}, low-resource $en\rightarrow x$ accuracy drops from 62.4\% on low-difficulty problems to 2.0\% on medium-difficulty problems and 1.2\% on high-difficulty problems; for \texttt{Qwen3-8B}, it drops from 68.0\% to 13.2\% and 3.2\%.
Thus, the low-resource trace-side problem is visible even on easier problems, and approaches collapse on medium- and high-difficulty problems.\footnote{We additionally evaluate an English-reasoning + target-language-output baseline and discuss the results in Appendix~\ref{app:english-reasoning-baseline}.}

\textbf{The accuracy gap also appears beyond the Qwen3 family.}
We further evaluate two distilled reasoning models on medium-difficulty PolyMath problems.
As shown in Table~\ref{tab:generalization_acc}, \texttt{DeepSeek-\allowbreak R1-\allowbreak Distill-\allowbreak Llama-8B} and \texttt{DeepSeek-\allowbreak R1-\allowbreak Distill-\allowbreak Qwen-32B} exhibit a similar pattern to \texttt{Qwen3-8B}: $en\rightarrow x$ performance decreases as language resource availability decreases, with the largest gaps observed for low-resource languages (Swahili and Telugu).
These results suggest that the trace-side degradation identified in our main experiments extends beyond the Qwen3 family and remains visible across different backbones and model scales.

\paragraph{Takeaway.}
The direction results separate two sources of reasoning gaps.
Non-English input can reduce accuracy, but the reasoning-trace language also has a large effect.
This effect is most severe when medium- and high-difficulty problems require low-resource reasoning traces.
This motivates a trace-side diagnosis: We next examine whether target-language reasoning traces preserve mathematical structures needed for a valid derivation.

\section{Directed Acyclic Trace Graph}
\seclabel{datg}

The results in \secref{accuracy} show that changing the reasoning language can sharply reduce performance even when the input problem remains in English. 
However, accuracy only tells us that target-language reasoning fails; it does not reveal how the generated reasoning trace breaks.
To diagnose these failures, we introduce \textbf{D}irected \textbf{A}cyclic \textbf{T}race \textbf{G}raph (\textbf{\datg}), a reference-conditioned post-hoc diagnostic framework (illustrated in Figure~\ref{fig:overview}) that compares reasoning execution across target languages through language-independent mathematical \emph{anchors} and dependency \emph{edges}.

\subsection{Reference Derivations and DAG Construction}
\seclabel{datg-construct}

\paragraph{Reference Derivations.} 

For each problem, we use \texttt{Qwen3.6-Plus} \citep{alibaba2026qwen36plus} to generate five English reference derivations for DAG construction.
Final-answer accuracy is evaluated separately against the official PolyMath answer using \texttt{Math-Verify}.
We use multiple reference derivations 
because a correct solution may choose a different route, introduce different intermediate variables, or order equivalent transformations differently.
This design broadens the set of accepted solutions and reduces the chance of penalizing a target-language reasoning trace for following a valid path that differs from a single English reference.\footnote{We quantify reference diversity by comparing the English references in Appendix~\ref{app:reference-diversity}.}

\paragraph{DAG Construction.} 
We then use \texttt{GPT-5.4} \citep{openai2026gpt54} to convert each reference derivation into a directed acyclic graph (DAG).\!\footnote{The trace-to-DAG prompt is provided in Appendix~\ref{app:dag-construction-prompt}.}
Each node in \datg represents a checkable mathematical anchor supported by the reference derivation.
Formally, we represent each node as a 4-tuple:
\setlength{\abovedisplayskip}{1.0pt} 
\setlength{\belowdisplayskip}{0.5pt}
\[
\mathrm{node} = (\textit{id}, \textit{anchor}, \textit{description}, \textit{parents}).
\]
Here, \textit{id} is a unique node identifier, \textit{anchor} stores the compact mathematical content of the node, \textit{description} provides a short English description, and \textit{parents} lists the identifiers of prerequisite nodes.
In practice, anchors serve one or more 
roles:\footnote{Appendix~\ref{app:node-semantics} gives additional role descriptions and concrete anchor examples used in DATG.}
\begin{itemize}[leftmargin=1.15em,itemsep=1pt,topsep=2pt]
    \item[] \textbf{Supporting facts}: quantities, rates, conditions, or target relations used by later steps.
    \item[] \textbf{Reasoning steps}: equations, computations, transformations, or intermediate states.
    \item[] \textbf{Answer-equivalent conclusions}: final mathematical states equivalent to the required answer.
\end{itemize}
\emph{Edges} are dependency-based rather than order-based.
A directed edge \(u\rightarrow v\) is created only when \(u\) is listed in the \textit{parents} field of \(v\), meaning that \(u\) is a prerequisite mathematical state for auditing or justifying \(v\).
Thus, independent steps are not connected merely because one appears earlier in the trace; for example, independent facts \(A\) and \(B\) may both point to \(C\) without an \(A\rightarrow B\) edge.


\subsection{Closed-Set Trace Alignment}
\seclabel{datg-align}

We use \texttt{GPT-5.4} \citep{openai2026gpt54} as an \emph{aligner} for closed-set trace alignment.\footnote{The
alignment prompt is provided in Appendix~\ref{app:closed-set-alignment-prompt}.}
Specifically, given a target-language reasoning trace and a reference DAG, the aligner will map evidence in the trace to anchors 
defined in the reference DAG; 
For each reference anchor, 
one of four statuses is assigned:
\begin{itemize}[leftmargin=1.15em,itemsep=1pt,topsep=2pt]
    \item[] \textsc{Commit}: the target-language reasoning trace correctly establishes the anchor.
    \item[] \textsc{Attempt}: the trace partially attempts the anchor but does not fully establish it.
    \item[] \textsc{Error}: the trace attempts the anchor but derives an incompatible mathematical state.
    \item[] \textsc{Missing}: the trace contains no relevant evidence for the anchor.
\end{itemize}
For example, for the anchor $r = 16 - 3 - 4 = 9$, a trace that derives $r=9$ receives \textsc{Commit}; a trace that only writes $r=16-3$ receives \textsc{Attempt}; and a trace that computes $r=11$ receives \textsc{Error}.
This closed-set design prevents fluent but irrelevant target-language text from receiving credit unless it supports a reference anchor.

\subsection{Metrics}
\seclabel{datg-metrics}

Let $G_i=(V_i,E_i)$ denote $i$-th reference DAG for a problem, where $V_i$ is the set of anchors and $E_i$ is the set of dependency edges.
For a target-language reasoning trace $T$, 
the alignment assigns each anchor $v\in V_i$ a status in \(\{\textsc{Commit}, \textsc{Attempt}, \textsc{Error}, \textsc{Missing}\}\).
We write $z_i(v)=\textsc{Commit}$ when $T$ correctly establishes anchor $v$ under alignment to $G_i$.

\paragraph{Committed Anchor Recall.}
Committed Anchor Recall (CAR) measures how many reference anchors are \emph{correctly} established by the target trace:
\[
\mathrm{CAR}(T,G_i) =
\frac{
\sum_{v\in V_i}
\mathbf{1}\{z_i(v)=\textsc{Commit}\}
}{
|V_i|
}.
\]
CAR captures whether the trace covers the mathematical content represented in the reference DAG.

\paragraph{Path Monotonic Fidelity.}
Path Monotonic Fidelity (PMF) measures whether the committed anchors preserve the dependency order of the reference DAG. 
Let $M(T,G_i)$ be the set of edges $(u,v)\in E_i$ whose endpoints are both committed and whose evidence order is dependency-compatible: the commit evidence for $u$ appears no later than that for $v$, or both commits fall in the same trace block.
We compute
\[
\mathrm{PMF}(T,G_i) = \frac{|M(T,G_i)|}{|E_i|}.
\]
PMF is coverage-aware because its denominator is the \emph{full} reference edge set \(E_i\), not only the edges between committed anchors.
Thus, missing anchors also penalize PMF through their incident edges, and therefore we interpret PMF together with CAR as dependency-preserving coverage. 

\begin{table*}[t]
\centering
\setlength{\belowcaptionskip}{0.5cm}
\scriptsize
\setlength{\tabcolsep}{2.6pt}
\renewcommand{\arraystretch}{0.92}
\resizebox{\textwidth}{!}{%
\begin{tabular}{llccccccccc}
\toprule
\multirow{2}{*}{Type} & \multirow{2}{*}{Lang.} &
\multicolumn{3}{c}{CAR $\uparrow$} &
\multicolumn{3}{c}{PMF $\uparrow$} &
\multicolumn{3}{c}{HAR $\downarrow$} \\
\cmidrule(lr){3-5}\cmidrule(lr){6-8}\cmidrule(lr){9-11}
 & & Low & Med. & High & Low & Med. & High & Low & Med. & High \\
\midrule
HRL & en
& .94/.98/.96 & .67/.78/.78 & .47/.61/.59
& .90/.96/.91 & .51/.68/.67 & .31/.46/.45
& .03/.01/.01 & .09/.04/.04 & .14/.08/.07 \\

HRL & fr
& .90/.96/.97 & .53/.74/.72 & .33/.52/.49
& .83/.93/.94 & .38/.61/.60 & .17/.39/.35
& .05/.02/.02 & .17/.07/.05 & .24/.12/.16 \\

MRL & th
& .89/.96/.95 & .46/.67/.67 & .29/.42/.42
& .79/.94/.91 & .29/.55/.54 & .15/.27/.28
& .06/.02/.02 & .27/.13/.10 & .30/.23/.23 \\

MRL & bn
& .61/.89/.94 & .26/.48/.53 & .13/.24/.29
& .45/.83/.90 & .11/.34/.37 & .03/.12/.15
& .26/.07/.04 & .45/.25/.19 & .51/.33/.31 \\

LRL & te
& .28/.75/.86 & .08/.19/.32 & .05/.10/.18
& .15/.67/.77 & .02/.06/.18 & .01/.03/.06
& .60/.18/.07 & .71/.47/.35 & .75/.56/.43 \\

LRL & sw
& .38/.59/.69 & .09/.12/.28 & .08/.06/.11
& .27/.50/.62 & .05/.04/.19 & .04/.01/.04
& .50/.34/.26 & .77/.69/.50 & .73/.68/.62 \\
\bottomrule
\end{tabular}}
\caption{
\datg results 
across model scales, with English included as the $en\rightarrow en$ reference
row (full per-language results in Appendix~\ref{app:datg-results}). 
HRL, MRL, and LRL denote high-, medium-, and low-resource language
groups.
Each cell reports \texttt{Qwen3-1.7B}/\texttt{Qwen3-4B}/\texttt{Qwen3-8B}.
CAR measures committed anchor coverage, PMF measures coverage-aware dependency
fidelity, and HAR measures harmful mathematical actions.}
\label{tab:datg_main}
\end{table*}

\paragraph{Harmful Action Rate.}
Harmful Action Rate (HAR) measures how often the target trace introduces harmful mathematical moves during closed-set alignment. 
Along with assigning evidence statuses to reference anchors, the aligner extracts mathematical actions in the target trace that are relevant to the selected reference DAG. 
We treat an action as \emph{judgeable} if it is aligned to a reference anchor, uses an aligned anchor as a prerequisite, or makes an explicit claim about a quantity or relation in the reference DAG.

The aligner marks a judgeable action as \emph{harmful} if it introduces an active contradiction, uses a wrong dependency, performs incorrect arithmetic, repeats an unproductive pseudo-step, or makes a degenerate mathematical transition. 
Corrected or explicitly retracted branches are not counted as harmful. 
Let $\mathcal{A}_{\mathrm{judge}}(T,G_i)$ denote the judgeable actions extracted during alignment between target trace $T$ and reference DAG $G_i$, and let $\mathcal{A}_{\mathrm{harm}}(T,G_i)\subseteq \mathcal{A}_{\mathrm{judge}}(T,G_i)$ denote the subset marked as harmful. 
We compute
\[
\mathrm{HAR}(T,G_i)=
\frac{
|\mathcal{A}_{\mathrm{harm}}(T,G_i)|
}{
\max\left(1,\left|\mathcal{A}_{\mathrm{judge}}(T,G_i)\right|\right)
}.
\]
Thus, HAR captures harmful mathematical behavior identified by the aligner, rather than simply measuring missing anchor coverage.

\paragraph{Human Evaluation of DATG Reliability.}
We conduct a 240-instance human evaluation to validate the reliability of automatic DATG extraction and scoring.
Reference DAG nodes and edges have 97.5\% and 96.5\% validity, automatic CAR/PMF closely track human-audited alignments (MAE .049/.048; Pearson .963/.955), and HAR reaches 90.1\% precision over harmful-action candidates. 
Details are provided in Appendix~\ref{app:human-eval}.

\begin{table}[t]
\centering
\vspace{-0.2em}
\setlength{\abovecaptionskip}{0.45em}
\setlength{\belowcaptionskip}{0.15cm}
\small
\setlength{\tabcolsep}{5.0pt}
\renewcommand{\arraystretch}{1.00}
\setlength{\aboverulesep}{0.45ex}
\setlength{\belowrulesep}{0.45ex}
\setlength{\cmidrulesep}{0.25ex}
\begin{tabular*}{\columnwidth}{@{\extracolsep{\fill}}lccc@{}}
\toprule

Difficulty & CAR (C/I) & PMF (C/I) & HAR (C/I) \\
\midrule
Low    & .96/.56 & .93/.40 & .03/.31 \\
Medium & .86/.40 & .75/.26 & .05/.29 \\
High   & .79/.26 & .67/.13 & .07/.34 \\
\midrule
All    & .91/.36 & .85/.22 & .04/.32 \\

\bottomrule
\end{tabular*}
\vspace{0.30em}
\caption{\datg scores under $en\rightarrow x$ stratified by final-answer correctness. C/I denotes correct/incorrect traces.}
\label{tab:datg_correctness}
\vspace{-0.3em}
\end{table}

\section{DATG Reveals Trace-Side Structural Degradation}
\seclabel{findings}

We now use \datg to inspect $en\rightarrow x$ traces from the Qwen3 models.
In this setting, the input problem
remains English and only the requested reasoning-trace language changes.
This keeps input understanding fixed and asks whether target-language
reasoning preserves the mathematical structure needed for execution.\footnote{We report \datg evaluation for the additional \texttt{DeepSeek-R1-Distill-Llama-8B} and \texttt{DeepSeek-R1-Distill-Qwen-32B} models in Appendix~\ref{app:cross-family-datg}.}

\subsection{Structural Degradation Across Languages}

Table~\ref{tab:datg_main} reports a subset of \datg-diagnosis results. 
Full per-language results
are in Appendix~\ref{app:datg-results}.


\textbf{Low-resource target-language traces show the sharpest structural
degradation.}
On medium-difficulty problems with \texttt{Qwen3-4B}, English reaches CAR
.78 and PMF .68 with HAR .04. French remains relatively close, with CAR
.74, PMF .61, and HAR .07. By contrast, Swahili drops to CAR .12 and
PMF .04 with HAR .69, and Telugu reaches only CAR .19 and PMF .06 with
HAR .47. Scaling to \texttt{Qwen3-8B} improves low-resource traces but does
not close the gap: Swahili rises to CAR .28 and PMF .19, still far below
English at CAR .78 and PMF .67. Thus, the low-resource accuracy gap is
accompanied by a concrete structural pattern: required anchors are less often
established, prerequisite relations are less often preserved, and harmful
mathematical actions are more frequent.

\textbf{Structural degradation is more severe on harder problems.}
For \texttt{Qwen3-4B}, English drops from CAR .98 and PMF .96 on
low-difficulty problems to CAR .61 and PMF .46 on high-difficulty problems,
with HAR increasing from .01 to .08. The drop is much sharper in
low-resource target languages. Swahili falls from CAR .59 and PMF .50 to
CAR .06 and PMF .01, while HAR rises from .34 to .68. Telugu similarly
falls from CAR .75 and PMF .67 to CAR .10 and PMF .03, with HAR
increasing from .18 to .56. This suggests that target-language reasoning
is especially fragile for longer derivations with more dependencies.

\subsection{Structural Scores Track Answer Correctness}

Since \datg-based metrics do not use final-answer correctness, we ask whether \emph{its structural metrics still distinguish successful from failed target-language reasoning traces}. 
Table~\ref{tab:datg_correctness} reports \datg scores for $en\rightarrow x$ traces averaged over target languages and model scales. 
Full model--difficulty--language results are provided in Appendix~\ref{app:datg-correctness-full}.

\textbf{Correct traces consistently show higher anchor coverage and dependency fidelity, while incorrect traces contain more harmful actions.} 
Across all difficulties, correct traces reach CAR .91 and PMF .85 with HAR .04, compared with CAR .36, PMF .22, and HAR .32 for incorrect traces. 
The separation remains large on high-difficulty problems: correct traces average CAR .79 and PMF .67, while incorrect traces fall to CAR .26 and PMF .13.
This shows that \datg is not merely restating aggregate accuracy differences; it identifies structural properties of reasoning traces that are strongly associated with final-answer success and failure.

\subsection{Failure Modes Exposed by DATG}


\begin{figure}[t]
  \centering
\setlength{\belowcaptionskip}{0.6cm}
\includegraphics[width=0.99\columnwidth]{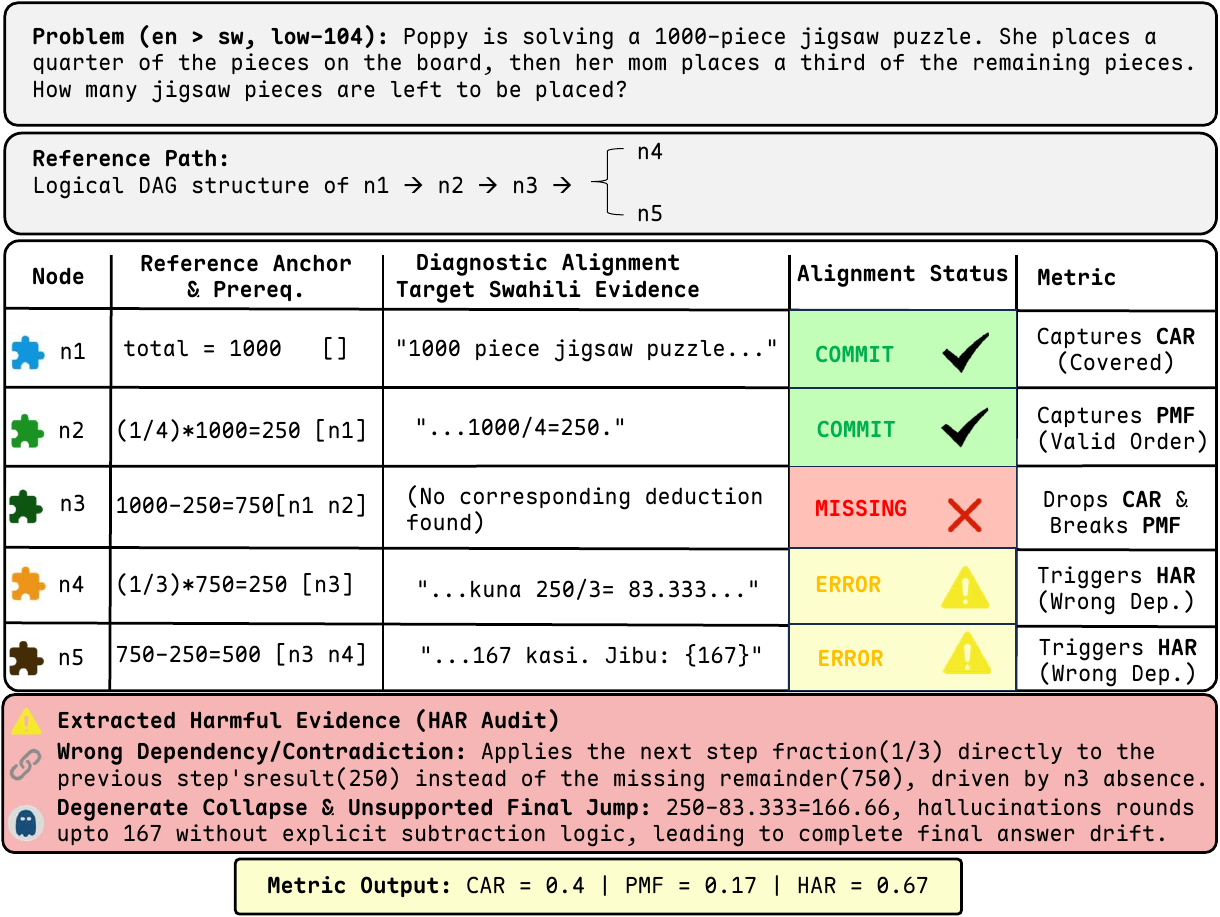}
  \caption{
Example of \datg alignment for an incorrect $en\rightarrow sw$ trace. The
Swahili trace commits the initial total and first fraction step, but misses the
required remainder anchor before applying the next fraction. The missing
prerequisite causes wrong dependency use and propagates to an incorrect final answer.
}
  \label{fig:case_study}
\end{figure}

\begin{table*}[t]
\centering
\setlength{\belowcaptionskip}{0.4cm}
\small
\setlength{\tabcolsep}{7pt}
\renewcommand{\arraystretch}{1.08}
\resizebox{\textwidth}{!}{%
\begin{tabular}{l ccc ccc ccc}
\toprule
\multirow{2}{*}{\textbf{Method}} & \multicolumn{3}{c}{\textbf{Accuracy (\%) $\uparrow$}} & \multicolumn{3}{c}{\textbf{Decoding Cost}} & \multicolumn{3}{c}{\textbf{DATG Diagnostic Metrics}} \\
\cmidrule(lr){2-4} \cmidrule(lr){5-7} \cmidrule(lr){8-10}
 & \textbf{sw} & \textbf{te} & \textbf{Avg} & \textbf{Tokens (M)} & \textbf{Retry (\%)} & \textbf{Trials} & \textbf{CAR $\uparrow$} & \textbf{PMF $\uparrow$} & \textbf{HAR $\downarrow$} \\
\midrule
\multicolumn{10}{l}{\textit{Targeted Test-Time Controls (Ours)}} \\
\quad \textbf{Loop-Retry} & \textbf{35.2} & 8.8 & \textbf{22.0} & 1.809 & 50.4 & 1.90 & .30 & \textbf{.18} & .51 \\
\quad \textbf{Formula-Retry} & 18.4 & 9.6 & 14.0 & 1.937 & 46.8 & 2.00 & \textbf{.30} & .13 & \textbf{.38} \\
\midrule
\multicolumn{10}{l}{\textit{Decoding \& Generic Retry Baselines}} \\
\quad Repetition Penalty & 9.6 & \textbf{12.8} & 11.2 & 1.623 & 0.0 & 1.00 & .19 & .09 & .60 \\
\quad Fixed Retry & 14.4 & 8.8 & 11.6 & 1.821 & 100.0 & 2.00 & .21 & .14 & .63 \\
\quad Best-of-2 & 8.0 & \textbf{12.8} & 10.4 & 3.543 & 100.0 & 2.00 & .18 & .08 & .62 \\
\quad Random Retry & 11.2 & 8.0 & 9.6 & 1.548 & 52.8 & 1.53 & .22 & .13 & .61 \\
\quad Base & 4.8 & 8.0 & 6.4 & 1.772 & 0.0 & 1.00 & .18 & .07 & .62 \\
\bottomrule
\end{tabular}}
\caption{Test-time control results on Swahili (\textbf{sw}) and Telugu (\textbf{te}) with \texttt{Qwen3-4B}. \textbf{Tokens (M)} denotes total decoded tokens excluding prompts. \textbf{Retry (\%)} is the percentage of examples triggering at least one retry, and \textbf{Trials} is the average attempts per example. CAR, PMF, and HAR are evaluated under the 5-DAG best-alignment protocol.}
\label{tab:control_main}
\end{table*}

Figure~\ref{fig:case_study} illustrates how \datg turns a target-language trace
into auditable structural decisions. The trace correctly commits the initial
total, $1000$, and the first placed amount, $1000/4=250$. However, it misses
the required remainder anchor, $1000-250=750$, before applying the next
fraction. The trace then applies the one-third operation to $250$ rather than
to the remaining $750$, producing an incompatible intermediate state and
propagating the error to the final answer.

This example shows why the metrics are complementary: CAR captures the missing
remainder anchor, PMF captures the broken prerequisite relation, and HAR captures
the harmful downstream use of the wrong quantity. These diagnostics motivate the next section's actionability check: whether lightweight test-time controls can improve low-resource reasoning by stabilizing trace execution.

\section{Test-Time Controls}
\seclabel{control}

The \datg diagnosis suggests that: if missing anchors, broken dependencies, and harmful mathematical actions are part of the low-resource failure pattern, then lightweight controls that stabilize trace execution should recover some accuracy.
We test this idea with \texttt{Qwen3-4B} in two low-resource languages: Swahili and Telugu on PolyMath medium-difficulty problems.\footnote{Appendix Table~\ref{tab:control_all_difficulty} reports Base, Loop-Retry, and Formula-Retry results on problems of all difficulties.}
We propose two controls: \emph{Loop-Retry}, which stops and resamples a trace when early repetition is detected, and \emph{Formula-Retry}, which provides a 
symbolic scaffold 
before the model continues reasoning in the target language. See Appendix~\ref{app:controlled-execution} for details.

\paragraph{Loop-Retry.}
Loop-Retry resamples traces that show early repetitive degeneration.
At a checkpoint placed at one quarter of the difficulty-dependent token budget, we inspect the most recent 256 generated tokens. 
If repetition statistics and type-token ratio indicate a loop, the current continuation is discarded and resampled under the same problem and target-language instruction.

\paragraph{Formula-Retry.}
Formula-Retry provides a compact answer-free symbolic scaffold before target-language reasoning continues, containing known quantities, variable relations, and formula templates but no final numeric result, boxed answer, or final-answer sentence.

\paragraph{Baselines.}
We compare against several simple test-time control methods to distinguish gains from targeted failure handling versus generic additional decoding.
\textbf{Repetition Penalty} uses a stronger repetition penalty throughout decoding without retries.
\textbf{Fixed Retry} always generates a second continuation for every example.
\textbf{Random Retry} retries examples selected uniformly at random, with the retry probability matched to the observed Loop-Retry retry rate for each language.
\textbf{Best-of-2} generates two independent continuations and reports an oracle upper-bound by selecting a correct continuation if either candidate matches the final answer; if neither continuation is correct, one incorrect continuation is retained for DATG analysis.

\subsection{Results and Discussion}

\paragraph{The proposed controls improve accuracy with limited additional decoding.}
Table~\ref{tab:control_main} reports results on medium-difficulty PolyMath. Both proposed controls, i.e., Loop-Retry and Formula-Retry, substantially improve over the Base setting and the other test-time control baselines. 
Loop-Retry achieves the highest average accuracy, increasing it from 6.4\% to 22.0\%, while Formula-Retry reaches 14.0\%. 
These gains require only modest additional decoding: Base uses 1.772M decoded tokens, compared with 1.809M for Loop-Retry and 1.937M for Formula-Retry.
In contrast, Best-of-2 decodes 3.543M tokens while reaching only 10.4\% average accuracy.
Other baselines, such as Repetition Penalty and Fixed Retry, use comparable or slightly fewer decoded tokens than our proposed methods, but yield substantially smaller accuracy improvements.
Thus, the gains are not simply a consequence of generating more tokens, but suggest that selectively stabilizing problematic traces makes more effective use of additional decoding.

\paragraph{Trace-side degradation can be partially mitigated.}
Re-applying \datg to the generated reasoning traces shows corresponding improvements in reasoning structure.
Relative to Base (CAR .18, PMF .07, HAR .62), Loop-Retry improves all three \datg metrics, reaching CAR .30, PMF .18, and HAR .51, consistent with better preservation of the required reasoning structure.
Formula-Retry achieves the highest CAR (.30) and lowest HAR (.38), indicating greater coverage of required mathematical anchors and fewer harmful mathematical actions.
In contrast, the generic test-time control baselines yield smaller and less consistent structural improvements over Base.
Together, these results show that the accuracy gains are accompanied by improvements in the trace-level properties identified by \datg, providing further evidence that the proposed test-time controls can mitigate low-resource reasoning degradation.

\section{Conclusion}

This paper shows that multilingual mathematical reasoning gaps arise not only from input-side understanding failures, but also from trace-side reasoning execution failures. 
By separating the input language from the requested reasoning language, we find that accuracy can drop sharply even when the problem remains in English. 
To diagnose how these failures occur, we introduce \datg, which compares reasoning traces through language-independent mathematical anchors and dependencies. 
We show that low-resource reasoning traces usually cover fewer anchors, preserve fewer dependency edges, and introduce more harmful mathematical actions. 
Finally, simple test-time controls partially recover accuracy, suggesting that future multilingual reasoning evaluation should look beyond final answers to the structure, ordering, and stability of generated reasoning traces.




\section*{Limitations}

Our $en\rightarrow x$ setting reduces the most direct input-side understanding confound by keeping the problem in English while varying the reasoning language. 
However, this does not fully eliminate all language-control issues. 
Although language drift is generally low for most target languages, it remains non-negligible for some language-model pairs. 

\datg also relies on strong commercial models for reference generation, DAG construction, and closed-set trace alignment. 
Although our expert manual audit supports its use as a trace-grounded aggregate diagnostic, \datg should not be interpreted as a formal proof verifier or as a routine metric.
Finally, our test-time controls are more diagnostic interventions rather than complete deployment solutions. 
Loop-Retry only addresses failures that appear as detectable repetition and cannot repair fluent but mathematically incorrect traces. 
Formula-Retry tests whether an answer-free symbolic scaffold can stabilize target-language reasoning execution, but automatically obtaining such scaffolds reliably remains an open problem.

\section*{Ethical Considerations}

\paragraph{Use of AI Assistants.}
The authors used \texttt{ChatGPT} for language editing, including grammar correction, clarity improvements, and coherence polishing, as well as limited assistance with code implementation.\footnote{\url{https://chatgpt.com/}}
All technical contributions, experimental design decisions, analyses, and final paper content remain the responsibility of the authors.

\section*{Acknowledgments}

This research was supported by the Munich Center
for Machine Learning (MCML) and German Research Foundation (DFG, grant SCHU 2246/14-1).

\bibliography{custom}

\FloatBarrier

\appendix
\captionsetup{hypcap=false}
\definecolor{PromptFrame}{HTML}{D8DEE8}
\definecolor{PromptHeader}{HTML}{E5E7EB}
\lstdefinestyle{promptstyle}{
  basicstyle=\ttfamily\scriptsize,
  breaklines=true,
  breakatwhitespace=true,
  columns=fullflexible,
  keepspaces=true,
  frame=single,
  framerule=0.3pt,
  framesep=3pt,
  rulecolor=\color{PromptFrame},
  backgroundcolor=\color{white},
  xleftmargin=0mm,
  xrightmargin=0mm,
  aboveskip=1pt,
  belowskip=3pt
}
\lstdefinestyle{examplestyle}{
  style=promptstyle,
  basicstyle=\ttfamily\tiny,
  belowskip=4pt
}
\providecommand{\promptheader}[2]{%
  \par\vspace{0.75em}\noindent
  \fcolorbox{PromptFrame}{PromptHeader}{%
    \makebox[\dimexpr\linewidth-2\fboxsep-2\fboxrule\relax][l]{%
      \strut\textcolor{black}{\bfseries #1}}}%
  \vspace{0.2pt}%
  \phantomsection\label{#2}%
}

\section{Prompt Templates and Language Control}
\label{app:prompts}

\subsection{Direction Probing Prompts}
\label{app:direction-prompts}

We evaluate four input--reasoning language settings by independently varying
the problem language and the requested reasoning-trace language. Here,
\emph{reasoning trace} refers to the model's generated CoT after the opening
thinking tag. Each query consists of three parts: a language-specific system
message, the problem statement as the user message, and a language-specific
assistant-side direct-first prefix.

\begin{center}
\centering
\footnotesize
\setlength{\tabcolsep}{3pt}
\renewcommand{\arraystretch}{1.08}
\begin{tabularx}{\columnwidth}{lY}
\toprule
Part & Content \\
\midrule
System & Language-specific solver instruction in the requested reasoning language. \\
User & Problem statement in English or the target language, depending on the direction. \\
Assistant & Direct-first prefix that starts the reasoning trace in the requested reasoning language. \\
\bottomrule
\end{tabularx}
\end{center}

\begin{center}
\footnotesize
\setlength{\tabcolsep}{3pt}
\renewcommand{\arraystretch}{1.08}
\begin{tabularx}{\columnwidth}{lYY}
\toprule
Setting & Problem statement & Prompt language \\
\midrule
$en\rightarrow en$ & English & English \\
$en\rightarrow x$  & English & Target language $x$ \\
$x\rightarrow en$  & Target language $x$ & English \\
$x\rightarrow x$   & Target language $x$ & Target language $x$ \\
\bottomrule
\end{tabularx}
\end{center}

In all four settings, the system message and assistant prefix use the requested
reasoning-trace language, while the user message supplies the problem in the
selected input language. This keeps input language and reasoning trace
language separable without changing the rest of the query format.

\subsection{Language-Specific Prompts}
\label{app:language-prompts}

\begin{figure*}[t]
\centering
\includegraphics[width=0.96\textwidth]{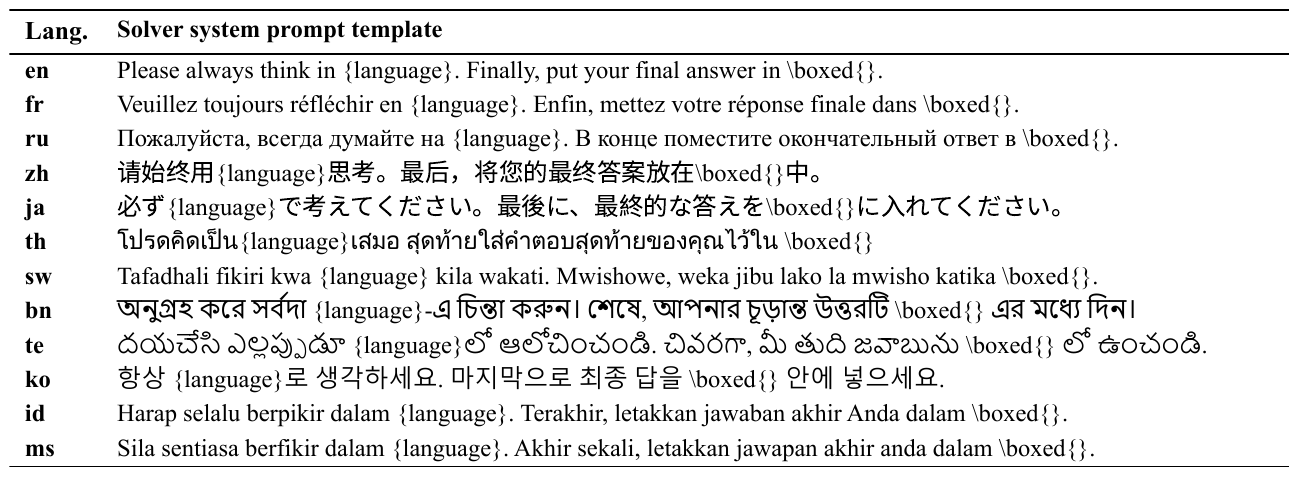}
\vspace{0.35em}
\IfFileExists{Picture_Table/language_direct_first_prefixes_cropped.pdf}{%
\includegraphics[width=0.96\textwidth]{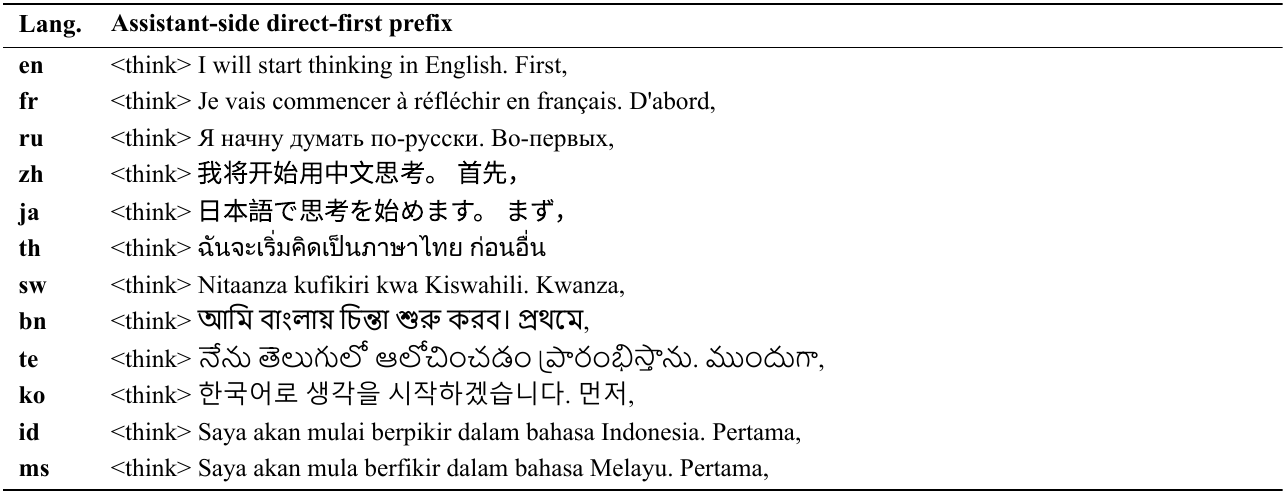}%
}{%
\includegraphics[width=0.96\textwidth]{Picture_Table/language_direct_first_prefixes.pdf}%
}
\caption{Language-specific solver system prompts and assistant-side direct-first prefixes.}
\label{fig:app_language_prompts}
\vspace{0.55em}
\scriptsize
\setlength{\tabcolsep}{4pt}
\renewcommand{\arraystretch}{1.03}
\begin{tabular}{lccc ccc ccc}
\toprule
\multirow{2}{*}{Pair}
& \multicolumn{3}{c}{1.7B}
& \multicolumn{3}{c}{4B}
& \multicolumn{3}{c}{8B} \\
\cmidrule(lr){2-4}\cmidrule(lr){5-7}\cmidrule(lr){8-10}
& Low & Med. & High & Low & Med. & High & Low & Med. & High \\
\midrule
en-fr & 98.4 & 93.6 & 95.2 & 99.2 & 100.0 & 99.2 & 89.6 & 75.2 & 79.2 \\
en-ru & 97.6 & 99.2 & 98.4 & 100.0 & 99.2 & 100.0 & 99.2 & 99.2 & 98.4 \\
en-zh & 84.8 & 84.0 & 92.0 & 85.6 & 84.0 & 95.2 & 85.6 & 82.4 & 96.8 \\
en-ja & 97.6 & 99.2 & 98.4 & 100.0 & 97.6 & 100.0 & 98.4 & 97.6 & 98.4 \\
en-th & 100.0 & 100.0 & 98.4 & 100.0 & 100.0 & 99.2 & 99.2 & 98.4 & 98.4 \\
en-sw & 93.6 & 94.4 & 93.6 & 96.0 & 96.8 & 96.8 & 96.8 & 84.8 & 92.8 \\
en-bn & 98.4 & 99.2 & 100.0 & 100.0 & 98.4 & 99.2 & 100.0 & 99.2 & 99.2 \\
en-te & 98.4 & 100.0 & 97.6 & 100.0 & 100.0 & 100.0 & 100.0 & 100.0 & 99.2 \\
en-ko & 92.0 & 85.6 & 86.4 & 96.8 & 84.0 & 91.2 & 93.6 & 81.6 & 84.0 \\
en-id & 98.4 & 98.4 & 99.2 & 91.2 & 93.6 & 92.8 & 95.2 & 78.4 & 72.8 \\
en-ms & 99.2 & 96.8 & 98.4 & 93.6 & 78.4 & 79.2 & 96.0 & 61.6 & 56.8 \\
\bottomrule
\end{tabular}
\captionof{table}{Reasoning language compliance under \(en\rightarrow x\). Values are percentages; each cell contains 125 traces.}
\label{tab:app_cot_compliance}
\end{figure*}

Figure~\ref{fig:app_language_prompts} lists the exact multilingual system
prompt templates and assistant-side direct-first prefixes used in the direction
probes. The prompt tables are rendered as vector PDFs to avoid Unicode font
issues under \texttt{pdflatex}.

\subsection{Reasoning Language Compliance}
\label{app:cot-compliance}

We report reasoning language consistency using the PolyMath-compatible language
consistency protocol. We evaluate only the generated CoT/reasoning trace under
\(en\rightarrow x\), not the final answer. For each trace, we remove
mathematical spans from \texttt{think\_trace}, including display and inline
math delimited by \texttt{\$\$...\$\$}, \texttt{\textbackslash(...\textbackslash)},
\texttt{\textbackslash[...\textbackslash]}, and \texttt{\$...\$}. We then apply
\texttt{langdetect.detect\_langs}. A trace is counted as language-consistent
when the detector returns a single language matching the requested reasoning
language. Following PolyMath, Chinese is mapped to \texttt{zh-cn}. Because
\texttt{langdetect} systematically maps Malay text to Indonesian, we
pre-specify a Malay--Indonesian alias: \texttt{ms} accepts \{\texttt{ms},
\texttt{id}\}, and \texttt{id} accepts \{\texttt{id}, \texttt{ms}\}.
Language consistency is reported separately from mathematical accuracy and is
not used to mark an answer incorrect. Each cell in
Table~\ref{tab:app_cot_compliance} is computed from 125 samples.

\section{Decoding and Generation Details}
\label{app:generation-details}

For the tested-model traces used in direction probing and DATG diagnosis, we
use a difficulty-specific maximum generation budget:
\[
\begin{aligned}
B_{\text{low}}&=4096,\\
B_{\text{med}}&=8192,\\
B_{\text{high}}&=16384.
\end{aligned}
\]
Across languages, directions, and difficulty levels, reasoning continuations
use the same decoding configuration: temperature \(0.6\), top-\(p=0.95\), and
the corresponding maximum budget \(B\). This keeps the direction comparisons
focused on input and reasoning language rather than on decoding variation.

\section{DATG Construction and Alignment Prompts}
\label{app:datg-prompts}

\subsection{Node and Edge Semantics}
\label{app:node-semantics}

A DATG node is a trace-supported, language-agnostic mathematical state that can
be audited against a reasoning trace. Nodes are not free-form sentences: their
\texttt{anchor} fields use equations, inequalities, assignments, set relations,
or controlled predicates. The \texttt{description} field is a short readability
gloss; it is not used as a string-matching target or as an independent scoring
field.

\begin{itemize}[leftmargin=*,itemsep=1pt,topsep=2pt]
    \item[-] \textbf{Given or fact relation}: an immutable quantity,
    condition, rate, or target relation from the problem or trace that is
    needed to audit later steps; e.g., \(eggs\_day=16\).
    \item[-] \textbf{Equation or setup relation}: an algebraic relation that
    defines how quantities are connected before or during computation; e.g.,
    \(remaining=eggs\_day-eaten-muffins\).
    \item[-] \textbf{Arithmetic transformation}: an explicit computation or
    simplification that changes one expression into an evaluated form; e.g.,
    \(16-3-4=9\).
    \item[-] \textbf{Derived intermediate result}: a computed value or
    symbolic state used by downstream reasoning; e.g., \(remaining=9\).
    \item[-] \textbf{Constraint, set, or predicate relation}: a checkable
    condition such as divisibility, membership, inequality, recurrence, or
    feasibility; e.g., \(\mathrm{Div}(56)\cap\{d:d>16\}=\{28,56\}\).
    \item[-] \textbf{Case, bound, or witness state}: a trace-stated case
    condition, extremal bound, candidate construction, or witness evaluation;
    e.g., \(x_{76}-x_{16}\le 41/800\).
    \item[-] \textbf{Contradiction or feasibility conclusion}: a controlled
    mathematical conclusion that a branch is impossible or satisfies a stated
    condition; e.g., \(bad\_coloring(k)\rightarrow contradiction\).
    \item[-] \textbf{Final answer-equivalent state}: the terminal mathematical
    state in the requested answer form; e.g., \(m+n=841\).
\end{itemize}

We do not create nodes for boxed-answer markers, output-format artifacts,
standalone constants, or purely verbal restatements that do not define a
checkable mathematical state. A DAG edge is defined by a node's
\texttt{parents}: each parent is a prerequisite mathematical state for the
child. These parent-defined edges are the dependency relations used by PMF.

\subsection{DAG Construction Prompt}
\label{app:dag-construction-prompt}

For each problem, the reference set contains five English reference derivations generated by \texttt{Qwen3.6-Plus}. These references are used to construct trace-grounded DAGs of language-agnostic mathematical anchors; final-answer accuracy is evaluated separately against the official PolyMath answer.

\promptheader{DAG construction prompt}{lst:dag_prompt}
\begin{lstlisting}[style=promptstyle]
System message:
You are a logic-faithful mathematical trace extractor.
Your task is to convert a verified Chain-of-Thought (CoT)
into a Directed Acyclic Graph (DAG) of language-agnostic
mathematical reasoning anchors.

CORE PRINCIPLE (TRACE-GROUNDED, MINIMAL-SUFFICIENT
REASONING GRAPH):
The DAG should represent the core reasoning steps plus the
minimal set of trace-supported input facts needed to make
those steps auditable. Each node must be anchored to a
language-neutral mathematical logic state supported by the
trace. Prefer computation anchors such as equations,
arithmetic transforms, derived intermediate results, and
final numeric conclusions that already appear in the trace.
However, do not delete a trace-stated quantity, rate,
condition, or target relation if a later calculation depends
on it and the dependency would become unclear without that
fact node. Do not invent a new symbolic system, solve the
problem again, or replace the trace with a cleaner
derivation unless the trace itself uses it.

NODE SCHEMA:
- node_id: n1, n2, ...
- anchor: one compact symbolic or controlled mathematical
  anchor, not a natural-language sentence
- description: a very short English gloss for readability only
- parents: prerequisite node_ids

Anchor forms are descriptive and non-mutually-exclusive. An
anchor may be a given fact, equation, arithmetic
transformation, derived intermediate result, necessary
relation, or answer-equivalent conclusion. Do not output an
anchor_type field.

SYMBOLIC ANCHOR RULE:
The anchor field must be language-independent in
serialization: use equations, inequalities, set relations,
assignments, or controlled predicates. Use compact variable
names such as eggs_day, remaining, price, S_+, D_i(k), or
target. Put all English prose in description. Do not write
natural English clauses in anchor such as "The question
asks...", "The condition means...", "there are...", or
"we get...".

Examples of acceptable anchors:
eggs_day = 16
remaining = eggs_day - eaten - muffins
16 - 3 - 4 = 9
Div(56) cap {d:d>16} = {28,56}
max(x_{76}-x_{16}) = 41/800
m+n = 841
bad_coloring(k) -> contradiction

Examples of unacceptable anchors:
Janet's ducks lay 16 eggs per day.
The divisors of 56 greater than 16 are 28 and 56.
The question asks how far he is from home.
The greatest value that x_76-x_16 can achieve is 41/800.

TRACE-FAITHFUL EXTREMAL CLAIMS:
For maximum, minimum, supremum, infimum, best/worst, or
sharp-bound problems, the goal is faithful extraction rather
than proof completion. Preserve the trace's explicit bound,
construction, witness, candidate value, limiting argument,
or trace-stated final conclusion as it appears. Do not
invent missing inequalities, missing constructions, missing
optimality arguments, or a cleaner proof. If the trace states
an extremal conclusion without spelling out every proof
component, keep it as a trace-stated or asserted conclusion
rather than rewriting it as a fully proven theorem.

FINAL NODE RULE:
The final_node_id must point to the trace's terminal
answer-equivalent mathematical state. Candidate values,
witness evaluations, constructions, and one-sided bounds
should normally be support nodes, not the final node, when
the trace later states a final answer or asserted conclusion.
For incomplete-but-correct traces, the terminal node may be
an asserted conclusion or trace-stated answer, but it must
correspond to the trace's final answer state rather than an
earlier partial result.

FINAL RESPONSE USE POLICY:
You may consult the final response only as a boundary cue
for the terminal answer-equivalent state. Do not use the
final response to invent missing intermediate reasoning or
to replace the trace's derivation. The DAG must still be
extracted from the reasoning trace, with the final response
used only to disambiguate the terminal answer when the
reasoning trace is clipped or contains multiple local
candidates.

MULTIPLE-PATH AND ENDPOINT-COUNT RULES:
If the trace contains two equivalent solution strategies,
extract one settled minimal path and drop alternate checks
that do not support the selected final node. If the trace
first derives a core count or intermediate expression and
later adds an endpoint, boundary case, or requested-form
conversion, keep the core count as support and make the
final node the later answer-equivalent conclusion.

FALSE STARTS AND ANSWER TARGETS:
Reasoning traces may contain false starts, abandoned
branches, exploratory guesses, or statements later corrected
by phrases such as "wait", "error", "actually",
"this failed", "not optimal", or "final check". Do not use
an abandoned intermediate conclusion as the final node.
Follow the trace's settled conclusion near the end. If the
trace converts an intermediate result into the requested
answer form, such as m+n, a residue, a count, a maximum
value, or a simplified expression, the final node must
include that requested answer-equivalent value.

RULES:
1) Graph must be acyclic.
2) Keep formulas as close as possible to the trace wording
   and arithmetic.
3) Do not rename variables unless the trace itself defines
   those variables.
4) If the trace uses plain arithmetic such as 16 - 7 = 9,
   keep that plain arithmetic instead of inventing notation
   like E_total.
5) Prefer computation anchors over verbal restatements of
   givens, but keep key immutable facts when they are needed
   to audit a later calculation.
6) Keep a given/fact node when it is a necessary dependency
   for later calculations and no downstream equation fully
   subsumes the same information with explicit numbers,
   units, and relation.
7) If the trace contains a combined equation such as
   2 + 1 = 3, prefer that combined equation over splitting it
   into separate result-only nodes.
8) Do not create nodes for boxed answers, final-answer
   markers, think tags, or other output-format artifacts.
9) Do not create standalone constant-only nodes unless they
   are unavoidable as trace-supported arithmetic anchors.
10) If a given/fact node must be kept, express it as a full
    relation from the trace rather than a bare literal.
11) If the trace includes both a setup form and a fully
    evaluated form for the same step, prefer the single more
    informative evaluated anchor unless the setup is needed
    as a separate dependency.
12) Preserve meaningful reasoning-path diversity across
    different traces, but within one DAG suppress purely
    notational, stylistic, formatting, verification, or
    alternate-solution branches.
13) Preserve the trace's logical granularity. Do not split
    or merge steps unless needed for DAG validity.
14) final_node_id must point to the final mathematical
    answer state or the last explicit answer-equivalent
    conclusion in the trace.
15) Follow the trace's final settled conclusion and ignore
    abandoned false-start conclusions.
16) For requested forms such as m+n, residues, counts, or
    simplified final expressions, make the final node contain
    that requested answer-equivalent form.
17) For extremal or sharp-bound conclusions, keep the trace's
    own proof status explicit in description, not as prose in
    anchor.
18) Do not hide uncertainty in the trace. If the trace
    contains a plausible but compressed conclusion, preserve
    it as a trace-stated conclusion; do not add a missing
    proof.
19) Output JSON only.

INTERNAL VALIDITY CHECK BEFORE OUTPUT:
- Emit nodes in dependency/topological order.
- Every retained node must support final_node_id directly or
  indirectly.
- final_node_id must be a terminal sink with no outgoing
  dependencies.
- If the trace evaluates the final expression, include the
  evaluated final value in the final anchor.
- Do this check silently; output only the final JSON object.

User message:

[Reasoning Trace to Parse]
<<<
{verified_reference_cot}
>>>

[Settled Trace Tail - final-node priority]
<<<TAIL
{reasoning_trace_tail}
TAIL>>>

[Final Response - answer boundary only]
<<<FINAL_RESPONSE
{final_response}
FINAL_RESPONSE>>>

[Task]
Extract one trace-faithful mathematical DAG and return
exactly one JSON object:

Output strict JSON only:
{
  "final_node_id": "nK",
  "nodes": [
    {
      "node_id": "n1",
      "anchor": "symbolic_or_controlled_math_state",
      "description": "short English gloss",
      "parents": []
    }
  ]
}

Hard constraints:
1. JSON only.
2. Do NOT output question_id or sample_id.
3. parents must reference only earlier nodes.
4. DAG must be acyclic and nodes must be emitted in
   dependency/topological order.
5. The DAG should capture a minimal-sufficient reasoning
   graph.
6. Each node must correspond to a mathematical logic anchor
   explicitly supported by the trace.
7. anchor must be symbolic or a controlled mathematical
   predicate, not an English sentence.
8. English prose is allowed only in description.
9. Reuse original formulas, arithmetic, and explicit numeric
   conclusions whenever possible.
10. If the trace states a fact in prose, normalize it into a
    compact relation.
11. Avoid paraphrasing formulas into newly invented symbolic
    notation.
12. Prefer computation anchors over verbal given/fact
    restatements.
13. Keep a given/fact node when it is needed as an input
    dependency.
14. If a given/fact node is kept, write it as a symbolic
    relation.
15. Prefer combined equations over split result-only nodes.
16. Prefer the single more informative evaluated anchor when
    possible.
17. Keep description very short.
18. Do NOT create nodes for \boxed{...}, Final Answer, think
    tags, or formatting artifacts.
19. Do NOT create standalone constant-only nodes unless
    unavoidable.
20. Preserve meaningful path differences across traces.
21. Ignore abandoned false starts and corrected conclusions.
22. final_node_id must point to a terminal sink node.
23. Requested answer forms such as m+n, residue, count, or
    simplified expression must be final.
24. Every node must support the final answer-equivalent node
    directly or indirectly.
25. If the trace evaluates the final expression, the final
    anchor must contain the evaluated final value.
26. If you add an evaluated final-answer node, update
    final_node_id to that node.
27. For extremal traces, do not require proof completion.
28. If the trace only evaluates a candidate or one side of a
    bound, preserve that limitation.
29. Faithfulness to the trace is more important than elegance
    or proof completeness.
30. Do not output anchor_type.
31. Before output, silently verify parent-first order, no
    cycles, all retained nodes reach final_node_id, and
    final_node_id is a terminal sink.

Budget guidance:
- target_max_nodes: {target_max_nodes}
- target_max_desc_chars_per_node: {target_max_desc_chars_per_node}
- target_max_pre_nodes_per_node: {target_max_pre_nodes_per_node}
\end{lstlisting}

\subsection{Closed-Set Alignment Prompt}
\label{app:closed-set-alignment-prompt}

For each target trace, the aligner maps trace evidence to a fixed reference DAG
in a closed set. It can only assign evidence to existing nodes and cannot
create new mathematical anchors.

\promptheader{Closed-set alignment prompt}{lst:align_prompt}
\begin{lstlisting}[style=promptstyle]
System message:
You are a closed-set mathematical reasoning extractor for
three metrics only: CAR (Committed anchor recall), PMF (path
monotonic fidelity), and HAR (harmful action rate).

Your job is extraction only, not scoring.

Main principles:
- Work in a closed set for node alignment: only map to the
  provided DAG nodes.
- Judge mathematical equivalence, not prose similarity.
- Symbol names may differ. Align by mathematical role,
  object identity, dependencies, and value, not by exact
  variable names.
- For invariant scalar quantities such as lengths, counts,
  ratios, divisibility conditions, and final arithmetic,
  COMMIT evidence must establish the same value or an
  algebraically identical form.
- For coordinate or auxiliary representation nodes,
  different coordinate conventions are allowed when the trace
  is internally consistent.
- Do not align nearby but different objects, auxiliary
  quantities, or generic intermediate equations to a DAG node
  merely because they occur in the same solution region.
- Return only the earliest useful events for each node.
- A node becomes activated only at its earliest valid COMMIT.
- If an early occurrence is partial or wrong, mark it ATTEMPT
  or ERROR.
- Do not use a later polished recap, final-answer summary, or
  post-hoc restatement as COMMIT evidence if an earlier
  equivalent computation already establishes the node.
- If a downstream computation correctly uses an intermediate
  value required by a parent node, that same local evidence
  may serve as the parent's COMMIT when no earlier explicit
  statement is available.
- Do not compute totals, rates, correctness labels, or final
  metric values.
- Do not segment the trace. Do not invent positions.

Allowed node event statuses:
- COMMIT: correctly establishes the DAG node.
- ATTEMPT: partially expressed or intended but not sufficient.
- ERROR: mathematically wrong derivation of that DAG node.

Strict COMMIT rule:
- The evidence must establish the node's anchor, not merely a
  related step in the same solution.
- If the evidence uses a different symbol or representation,
  require enough local context to identify the same
  mathematical role or object.
- A bare mention of a number, variable, expression, or example
  substitution is not COMMIT.
- Verbal mathematical statements in any language can be COMMIT
  when the same quantity/state/update is locally checkable.
- If the trace reaches the right final answer through an
  alternative path not represented by the DAG, leave unmatched
  DAG nodes empty rather than forcing a match.

For node events, provide only:
- evidence: exact short quote copied from the trace, <= 80
  characters.
- evidence_span: exact continuous quote copied from the
  trace, <= 120 characters.

For HAR, keep only local harmful mathematical actions:
- contradictory_steps: harmful contradictions, wrong
  dependency use, or wrong DAG-node attempts that remain
  active in the final reasoning state.
- harmful_loop_steps: harmful repeated math acts that revisit
  an established state without adding justified state.
- degenerate_steps: one short repeated/garbled pseudo-math
  unit if the trace collapses.

Do not output off-path branches, harmless redundancy lists,
verification steps, or benign restatements.
Do not copy long repeated paragraphs.
Each HAR list may contain at most 2 items, and
degenerate_steps at most 1 item.
If an ERROR node event can change the active reasoning state
or final answer, also include its same short evidence in the
appropriate HAR list.
Do not include a wrong intermediate branch in HAR if the
trace explicitly marks it as a mistake and replaces it before
using it for the final answer.
If uncertain, leave the node/list empty. Do not guess.
Output strict JSON only.

User message:
Target trace language: {target_lang}

[Reference Mathematical DAG]
{dag_json}

[Target Reasoning Trace]
<<<TRACE
{raw_trace}
TRACE>>>

Task:
Extract only what is needed for CAR, PMF, and HAR.
For each DAG node, return only the earliest useful
mathematically supported occurrences in chronological order.
This is a closed-set alignment task: only align to the
provided node inventory. Do not invent nodes.
A node is activated only by its earliest valid COMMIT. Later
re-checks or restatements are irrelevant and must not be
returned. Use ATTEMPT for partial or incomplete derivations,
ERROR for mathematically wrong derivations, and COMMIT only
for a correct and sufficient establishment of the node.
Variable names may differ; align by mathematical role, object
identity, dependencies, and value, not exact names.
If the trace reaches the right final answer through an
alternative path not represented by this DAG, leave unmatched
DAG nodes empty rather than forcing a match.
For HAR, return only local harmful mathematical actions in
contradictory_steps, harmful_loop_steps, or degenerate_steps.
If uncertain, leave the node empty and do not guess.
Do not estimate total_math_steps, effective_math_acts, or any
score field. Extraction only.

Return strict JSON only:
{
  "audit_results": {
    "n1":[{"status":"COMMIT","evidence":"...",
           "evidence_span":"..."}],
    "n4":[{"status":"ATTEMPT","evidence":"...",
           "evidence_span":"..."},
          {"status":"COMMIT","evidence":"...",
           "evidence_span":"..."}],
    "n5":[]
  },
  "contradictory_steps": [
    {"evidence":"...","evidence_span":"...",
     "category":"contradiction"}
  ],
  "harmful_loop_steps": [
    {"evidence":"...","evidence_span":"...",
     "category":"loop"}
  ],
  "degenerate_steps": [
    {"evidence":"...","category":"collapse"}
  ]
}
\end{lstlisting}

An empty list for a node in \texttt{audit\_results} represents a missing
anchor; the deterministic scorer maps such nodes to \textsc{Missing}. Thus
Missing is a scoring status rather than an aligner-emitted label.

For each target trace, we compute CAR, PMF, and HAR against all five candidate
reference DAGs. We select the most compatible alignment by maximizing CAR, then
PMF, and finally minimizing HAR. The LLM aligner performs closed-set semantic
alignment against the provided node inventory, using the \texttt{anchor} as the
primary mathematical target and \texttt{description} only as a readability gloss;
it is not asked to compute metric values or to string-match descriptions. The aligner returns status labels and \texttt{evidence\_span}s.
The scoring code then locates each span in the original trace to obtain
\texttt{start\_char}, \texttt{end\_char}, \texttt{commit\_progress}, and
\texttt{block\_id}. PMF is computed from the parent-defined DAG edges: for each
parent-to-child edge, the edge is ordered only if both endpoint nodes are
\textsc{COMMIT} and the parent commit progress is no later than the child
commit progress, or both commits fall in the same trace block. The final PMF
score is the number of ordered edges divided by the total number of DAG edges.

\section{Reference-Side Diversity}
\label{app:reference-diversity}

To quantify variation among the five English reference derivations for the same problem, we perform reference-to-reference alignment. For each English reference trace, we align it against the other four English reference DAGs for the same problem, exclude its own DAG, and select the most compatible ``best-other'' alignment by maximizing CAR, then PMF, and finally minimizing HAR. This measures whether different English references provide alternative derivation paths while preserving compatible core mathematical structure.

As shown in Table~\ref{tab:app_reference_diversity}, the overall best-other CAR reaches $0.98 \pm .08$ and PMF reaches $0.92 \pm .15$, with near-zero HAR ($0.00 \pm .02$). Even on high-difficulty problems, the derivations maintain high anchor coverage ($0.95$) and monotonic fidelity ($0.84$), confirming that the extracted reference DAGs share a mutually consistent mathematical foundation while allowing valid solution-path diversity.

\begin{table}[H]
\centering
\scriptsize
\setlength{\tabcolsep}{3.5pt}
\renewcommand{\arraystretch}{0.95}
\resizebox{\columnwidth}{!}{%
\begin{tabular}{lccc}
\toprule
Difficulty & Best-other CAR $\uparrow$ & Best-other PMF $\uparrow$ & Best-other HAR $\downarrow$ \\
\midrule
Low     & $1.00 \pm .00$ & $1.00 \pm .04$ & $.01 \pm .04$ \\
Medium  & $.99 \pm .04$ & $.93 \pm .12$ & $.00 \pm .00$ \\
High    & $.95 \pm .13$ & $.84 \pm .20$ & $.00 \pm .01$ \\
Overall & $.98 \pm .08$ & $.92 \pm .15$ & $.00 \pm .02$ \\
\bottomrule
\end{tabular}}
\caption{Reference-side diversity among English reference derivations used for DAG construction. Values are mean $\pm$ standard deviation over best-other reference-to-reference alignments.}
\label{tab:app_reference_diversity}
\end{table}

\FloatBarrier

\section{Human Evaluation of DATG}
\label{app:human-eval}

We use human evaluation as a diagnostic reliability check for DATG, not as a
benchmark for formal proof verification. Two annotators independently
audited reference DAG quality, closed-set trace alignment, and HAR candidates.
The expanded audit set covers French, Swahili, Bengali, and Telugu, with
60 samples per language. For each language, we sample 20 instances from each
difficulty level. These languages include both relatively stable
multilingual settings and the lower-resource settings where the automatic DATG
analysis shows the largest degradation.

Inter-annotator agreement is computed at the annotation layer rather than only
for the final aggregate scores. We report MAE and correlation for CAR/PMF
because these are continuous scores derived from committed-node alignments
rather than categorical labels.

Table~\ref{tab:app_human_eval} provides direct support for DATG as an aggregate,
trace-grounded diagnostic representation. The reference DAGs have consistently
high node and edge validity across all four languages, including Bengali and
Telugu, indicating that the extracted reference structures are mathematically
well-formed enough for aggregate diagnostic use.

The CAR and PMF columns show that the automatic scores track human-audited
alignments closely, with low MAE and strong Pearson/Spearman correlations in
the aggregate. This is the key calibration result for our use case: DATG is not
claimed to certify formal proof validity, but its closed-set anchor alignment
is reliable enough to support comparisons of coverage and dependency order
across languages, difficulties, and model scales. For HAR, the automatic
extractor produces 162 harmful-action candidates, of which 146 are confirmed by
human annotators, yielding a precision of 90.1\%. This supports HAR as an
aggregate signal for harmful mathematical actions.

DATG should therefore be read as reference-conditioned, trace-grounded
diagnosis rather than formal proof verification. It does not define a formal
proof language, proof obligations, or mechanically checkable inference rules.
PMF checks whether committed anchors appear in an order compatible with a
reference DAG; it does not prove that every child node is logically entailed by
its parents. The reference DAGs are LLM-extracted from reference derivations,
so they provide diagnostic structure, not sound proof certificates.

\begin{table}[H]
\centering
\scriptsize
\setlength{\tabcolsep}{1.5pt}
\renewcommand{\arraystretch}{0.95}
\begin{tabular*}{\columnwidth}{@{\extracolsep{\fill}}lrrrrcccc@{}}
\toprule
\multirow{2}{*}{Lang.} & \multirow{2}{*}{$n$} &
\multicolumn{2}{c}{DAG} &
\multicolumn{2}{c}{CAR} &
\multicolumn{2}{c}{PMF} &
\multirow{2}{*}{HAR} \\
\cmidrule(lr){3-4}\cmidrule(lr){5-6}\cmidrule(lr){7-8}
& & Node & Edge & MAE & $r$ & MAE & $r$ & \\
\midrule
fr  & 60  & 98.7 & 96.1 & .079 & .886/.871 & .118 & .846/.857 & 65.4 \\
sw  & 60  & 97.7 & 95.3 & .034 & .965/.935 & .014 & .985/.904 & 100.0 \\
bn  & 60  & 96.8 & 98.1 & .046 & .967/.943 & .043 & .970/.966 & 85.0 \\
te  & 60  & 96.8 & 96.4 & .036 & .969/.918 & .015 & .985/.954 & 97.7 \\
\midrule
All & 240 & 97.5 & 96.5 & .049 & .963/.955 & .048 & .955/.954 & 90.1 \\
\bottomrule
\end{tabular*}
\vspace{0.35em}

\begin{tabular*}{\columnwidth}{@{\extracolsep{\fill}}lcccc@{}}
\toprule
Agreement & Node & Edge & Node-event & HAR \\
\midrule
Pooled Cohen's $\kappa$ & .753 & .770 & .938 & .798 \\
\bottomrule
\end{tabular*}
\caption{Human evaluation of DATG. DAG validity is in percentages; CAR/PMF report MAE and Pearson/Spearman correlation against human-audited alignments. HAR precision is over 162 candidates, with all 240 double-annotated instances in the aggregate row.}
\label{tab:app_human_eval}
\end{table}

\clearpage
\begin{table*}[t]
\noindent\begin{minipage}[t]{\columnwidth}
  \section{Full Direction-Probing Accuracy}
  \label{app:direction-accuracy}
\end{minipage}%
\par\vspace{0.3em}
\centering
\scriptsize
\setlength{\tabcolsep}{2.15pt}
\renewcommand{\arraystretch}{0.98}
\resizebox{0.98\textwidth}{!}{%
\begin{tabular}{llrrrrrrrrrrrr}
\toprule
Difficulty & Direction & en & fr & ru & zh & ja & ko & id & ms & th & bn & sw & te \\
\midrule
\multicolumn{14}{l}{\textbf{1.7B}} \\
low & en-en & 81.6 & -- & -- & -- & -- & -- & -- & -- & -- & -- & -- & -- \\
low & en-x & -- & 73.6 & 75.2 & 77.6 & 67.2 & 74.4 & 71.2 & 58.4 & 72.0 & 44.8 & 27.2 & 7.2 \\
low & x-en & -- & 72.8 & 78.4 & 71.2 & 63.2 & 64.0 & 73.6 & 75.2 & 64.8 & 55.2 & 8.8 & 40.8 \\
low & x-x & -- & 63.2 & 68.8 & 74.4 & 59.2 & 60.8 & 67.2 & 61.6 & 64.8 & 34.4 & 4.0 & 13.6 \\
medium & en-en & 36.8 & -- & -- & -- & -- & -- & -- & -- & -- & -- & -- & -- \\
medium & en-x & -- & 31.2 & 32.0 & 36.8 & 18.4 & 17.6 & 24.8 & 16.8 & 22.4 & 5.6 & 1.6 & 0.8 \\
medium & x-en & -- & 38.4 & 39.2 & 40.8 & 36.0 & 28.8 & 37.6 & 35.2 & 31.2 & 30.4 & 21.6 & 24.8 \\
medium & x-x & -- & 24.0 & 32.8 & 36.8 & 25.6 & 14.4 & 25.6 & 26.4 & 14.4 & 7.2 & 1.6 & 0.0 \\
high & en-en & 23.2 & -- & -- & -- & -- & -- & -- & -- & -- & -- & -- & -- \\
high & en-x & -- & 11.2 & 8.8 & 19.2 & 10.4 & 5.6 & 9.6 & 5.6 & 8.0 & 0.0 & 3.2 & 0.8 \\
high & x-en & -- & 17.6 & 20.0 & 25.6 & 23.2 & 23.2 & 16.8 & 20.0 & 16.0 & 17.6 & 12.0 & 15.2 \\
high & x-x & -- & 13.6 & 13.6 & 20.0 & 6.4 & 8.0 & 8.8 & 10.4 & 4.8 & 1.6 & 0.8 & 0.0 \\
\midrule
\multicolumn{14}{l}{\textbf{4B}} \\
low & en-en & 92.8 & -- & -- & -- & -- & -- & -- & -- & -- & -- & -- & -- \\
low & en-x & -- & 88.8 & 89.6 & 91.2 & 91.2 & 89.6 & 92.0 & 92.0 & 90.4 & 84.0 & 52.8 & 72.0 \\
low & x-en & -- & 87.2 & 91.2 & 85.6 & 78.4 & 87.2 & 84.8 & 87.2 & 80.8 & 79.2 & 24.0 & 68.0 \\
low & x-x & -- & 83.2 & 88.8 & 84.8 & 76.8 & 80.8 & 86.4 & 85.6 & 78.4 & 76.0 & 24.0 & 49.6 \\
medium & en-en & 53.6 & -- & -- & -- & -- & -- & -- & -- & -- & -- & -- & -- \\
medium & en-x & -- & 43.2 & 51.2 & 49.6 & 41.6 & 36.8 & 47.2 & 44.8 & 35.2 & 16.0 & 0.0 & 4.0 \\
medium & x-en & -- & 36.8 & 36.8 & 31.2 & 36.0 & 34.4 & 40.8 & 40.8 & 39.2 & 32.8 & 25.6 & 32.8 \\
medium & x-x & -- & 32.8 & 32.0 & 24.8 & 28.8 & 31.2 & 33.6 & 31.2 & 29.6 & 16.0 & 2.4 & 2.4 \\
high & en-en & 41.6 & -- & -- & -- & -- & -- & -- & -- & -- & -- & -- & -- \\
high & en-x & -- & 31.2 & 32.8 & 36.0 & 23.2 & 24.0 & 30.4 & 26.4 & 16.8 & 10.4 & 0.8 & 1.6 \\
high & x-en & -- & 39.2 & 40.0 & 43.2 & 39.2 & 43.2 & 38.4 & 41.6 & 33.6 & 36.0 & 20.8 & 31.2 \\
high & x-x & -- & 36.0 & 37.6 & 42.4 & 24.8 & 22.4 & 25.6 & 29.6 & 19.2 & 10.4 & 0.8 & 3.2 \\
\midrule
\multicolumn{14}{l}{\textbf{8B}} \\
low & en-en & 94.4 & -- & -- & -- & -- & -- & -- & -- & -- & -- & -- & -- \\
low & en-x & -- & 92.0 & 94.4 & 92.0 & 91.2 & 87.2 & 90.4 & 92.0 & 92.0 & 87.2 & 59.2 & 76.8 \\
low & x-en & -- & 88.8 & 92.0 & 88.0 & 86.4 & 89.6 & 92.0 & 89.6 & 89.6 & 84.0 & 64.8 & 76.8 \\
low & x-x & -- & 86.4 & 92.0 & 88.8 & 80.0 & 84.0 & 88.0 & 88.0 & 89.6 & 80.8 & 23.2 & 58.4 \\
medium & en-en & 53.6 & -- & -- & -- & -- & -- & -- & -- & -- & -- & -- & -- \\
medium & en-x & -- & 50.4 & 49.6 & 54.4 & 39.2 & 36.0 & 48.0 & 48.0 & 48.8 & 34.4 & 16.0 & 10.4 \\
medium & x-en & -- & 34.4 & 34.4 & 36.8 & 33.6 & 48.8 & 52.0 & 49.6 & 28.0 & 41.6 & 35.2 & 48.8 \\
medium & x-x & -- & 47.2 & 45.6 & 48.0 & 46.4 & 40.0 & 50.4 & 47.2 & 43.2 & 26.4 & 5.6 & 12.8 \\
high & en-en & 40.8 & -- & -- & -- & -- & -- & -- & -- & -- & -- & -- & -- \\
high & en-x & -- & 29.6 & 36.8 & 39.2 & 22.4 & 18.4 & 28.0 & 27.2 & 25.6 & 12.8 & 2.4 & 4.0 \\
high & x-en & -- & 39.2 & 43.2 & 41.6 & 45.6 & 41.6 & 40.0 & 39.2 & 43.2 & 35.2 & 32.0 & 34.4 \\
high & x-x & -- & 33.6 & 37.6 & 39.2 & 32.0 & 24.0 & 30.4 & 35.2 & 27.2 & 12.0 & 4.0 & 4.0 \\
\bottomrule
\end{tabular}}
\caption{Full direction-probing accuracy by model, difficulty, direction, and language. Values are percentages; each non-empty cell is evaluated on 125 PolyMath problems.}
\label{tab:app_direction_accuracy_full}
\end{table*}
\clearpage

\begin{table*}[t]
\noindent\begin{minipage}[t]{\columnwidth}
  \section{Full DATG Results}
  \label{app:datg-results}
\end{minipage}%
\par\vspace{0.3em}
\centering
\scriptsize
\setlength{\tabcolsep}{2.6pt}
\renewcommand{\arraystretch}{0.72}
\resizebox{0.86\textwidth}{!}{%
\begin{tabular}{llccc ccc ccc}
\toprule
& & \multicolumn{3}{c}{CAR} & \multicolumn{3}{c}{PMF} & \multicolumn{3}{c}{HAR} \\
\cmidrule(lr){3-5}\cmidrule(lr){6-8}\cmidrule(lr){9-11}
Pair & Difficulty & 1.7B & 4B & 8B & 1.7B & 4B & 8B & 1.7B & 4B & 8B \\
\midrule
\multirow{3}{*}{en-en} & High & .47 & .61 & .59 & .31 & .46 & .45 & .14 & .08 & .07 \\
& Medium & .67 & .78 & .78 & .51 & .68 & .67 & .09 & .04 & .04 \\
& Low & .94 & .98 & .96 & .90 & .96 & .91 & .03 & .01 & .01 \\
\midrule
\multirow{3}{*}{en-fr} & High & .33 & .52 & .49 & .17 & .39 & .35 & .24 & .12 & .16 \\
& Medium & .53 & .74 & .72 & .38 & .61 & .60 & .17 & .07 & .05 \\
& Low & .90 & .96 & .97 & .83 & .93 & .94 & .05 & .02 & .02 \\
\midrule
\multirow{3}{*}{en-zh} & High & .43 & .59 & .54 & .27 & .45 & .40 & .17 & .09 & .09 \\
& Medium & .62 & .80 & .79 & .48 & .68 & .67 & .11 & .03 & .03 \\
& Low & .93 & .97 & .98 & .88 & .94 & .95 & .04 & .02 & .01 \\
\midrule
\multirow{3}{*}{en-ja} & High & .31 & .47 & .40 & .16 & .32 & .25 & .31 & .20 & .18 \\
& Medium & .51 & .73 & .60 & .34 & .61 & .43 & .23 & .09 & .13 \\
& Low & .89 & .98 & .96 & .80 & .97 & .93 & .08 & .01 & .02 \\
\midrule
\multirow{3}{*}{en-ru} & High & .37 & .55 & .55 & .22 & .40 & .40 & .24 & .12 & .11 \\
& Medium & .60 & .78 & .71 & .45 & .66 & .57 & .16 & .06 & .08 \\
& Low & .90 & .98 & .98 & .81 & .96 & .95 & .06 & .01 & .01 \\
\midrule
\multirow{3}{*}{en-ko} & High & .27 & .45 & .34 & .12 & .31 & .20 & .32 & .24 & .20 \\
& Medium & .49 & .70 & .57 & .32 & .55 & .41 & .23 & .11 & .12 \\
& Low & .88 & .97 & .93 & .79 & .95 & .87 & .08 & .02 & .03 \\
\midrule
\multirow{3}{*}{en-id} & High & .33 & .52 & .47 & .19 & .39 & .34 & .27 & .13 & .17 \\
& Medium & .53 & .76 & .69 & .36 & .64 & .57 & .18 & .05 & .12 \\
& Low & .89 & .97 & .95 & .82 & .94 & .90 & .06 & .02 & .02 \\
\midrule
\multirow{3}{*}{en-ms} & High & .31 & .50 & .45 & .16 & .34 & .30 & .27 & .15 & .17 \\
& Medium & .49 & .73 & .67 & .32 & .61 & .54 & .17 & .06 & .08 \\
& Low & .89 & .97 & .94 & .80 & .95 & .91 & .06 & .02 & .03 \\
\midrule
\multirow{3}{*}{en-th} & High & .29 & .42 & .42 & .15 & .27 & .28 & .30 & .23 & .23 \\
& Medium & .46 & .67 & .67 & .29 & .55 & .54 & .27 & .13 & .10 \\
& Low & .89 & .96 & .95 & .79 & .94 & .91 & .06 & .02 & .02 \\
\midrule
\multirow{3}{*}{en-bn} & High & .13 & .24 & .29 & .03 & .12 & .15 & .51 & .33 & .31 \\
& Medium & .26 & .48 & .53 & .11 & .34 & .37 & .45 & .25 & .19 \\
& Low & .61 & .89 & .94 & .45 & .83 & .90 & .26 & .07 & .04 \\
\midrule
\multirow{3}{*}{en-sw} & High & .08 & .06 & .11 & .04 & .01 & .04 & .73 & .68 & .62 \\
& Medium & .09 & .12 & .28 & .05 & .04 & .19 & .77 & .69 & .50 \\
& Low & .38 & .59 & .69 & .27 & .50 & .62 & .50 & .34 & .26 \\
\midrule
\multirow{3}{*}{en-te} & High & .05 & .10 & .18 & .01 & .03 & .06 & .75 & .56 & .43 \\
& Medium & .08 & .19 & .32 & .02 & .06 & .18 & .71 & .47 & .35 \\
& Low & .28 & .75 & .86 & .15 & .67 & .77 & .60 & .18 & .07 \\
\bottomrule
\end{tabular}}
\vspace{-0.5em}
\caption{Full DATG results by language pair, difficulty, model scale, and metric.}
\label{tab:app_datg_full}
\end{table*}
\clearpage

\clearpage
\twocolumn[{%
  \begin{minipage}[t]{\columnwidth}
    \section{Cross-Family Medium-Difficulty DATG Results}
    \label{app:cross-family-datg}
  \end{minipage}%
  \par\vspace{0.4em}
  \begin{center}
    \scriptsize
    \setlength{\tabcolsep}{4.0pt}
    \renewcommand{\arraystretch}{1.08}
    \begin{tabular*}{\textwidth}{@{\extracolsep{\fill}}llrrrrrrrr@{}}
    \toprule
    \multirow{2}{*}{Group} & \multirow{2}{*}{Pair}
    & \multicolumn{4}{c}{\texttt{DeepSeek-R1-Distill-Llama-8B}}
    & \multicolumn{4}{c}{\texttt{DeepSeek-R1-Distill-Qwen-32B}} \\
    \cmidrule(lr){3-6}\cmidrule(lr){7-10}
    & & Acc. (\%) & CAR & PMF & HAR & Acc. (\%) & CAR & PMF & HAR \\
    \midrule
    EN  & en$\rightarrow$en & 33.6 & .44 & .32 & .40 & 50.4 & .55 & .43 & .31 \\
    \midrule
    HRL & en$\rightarrow$fr & 24.0 & .45 & .28 & .37 & 47.2 & .55 & .42 & .29 \\
    HRL & en$\rightarrow$ru & 32.0 & .50 & .35 & .32 & 48.8 & .61 & .48 & .24 \\
    HRL & en$\rightarrow$zh & 36.0 & .58 & .45 & .27 & 54.4 & .72 & .57 & .17 \\
    HRL & en$\rightarrow$ja & 26.4 & .53 & .37 & .29 & 38.4 & .53 & .38 & .29 \\
    HRL & en$\rightarrow$ko & 31.2 & .54 & .40 & .31 & 43.2 & .55 & .40 & .28 \\
    \midrule
    MRL & en$\rightarrow$id & 27.2 & .42 & .26 & .35 & 36.0 & .57 & .41 & .24 \\
    MRL & en$\rightarrow$ms & 21.6 & .37 & .21 & .39 & 40.8 & .52 & .37 & .25 \\
    MRL & en$\rightarrow$th & 15.2 & .30 & .19 & .51 & 42.4 & .60 & .48 & .23 \\
    MRL & en$\rightarrow$bn & 7.2  & .12 & .04 & .74 & 36.8 & .54 & .39 & .30 \\
    \midrule
    LRL & en$\rightarrow$sw & 2.4  & .09 & .02 & .71 & 17.6 & .33 & .20 & .47 \\
    LRL & en$\rightarrow$te & 18.4 & .21 & .12 & .61 & 8.0  & .26 & .13 & .48 \\
    \bottomrule
    \end{tabular*}
    \captionof{table}{Medium-difficulty cross-family results on PolyMath. Accuracy uses the same final-answer scoring protocol and source as Table~\ref{tab:generalization_acc}; CAR/PMF/HAR are computed with the 5-DAG best-alignment protocol.}
    \label{tab:cross_family_datg}
    \vspace{0.8em}
  \end{center}
}]

Table~\ref{tab:cross_family_datg} reports medium-difficulty PolyMath results for the additional reasoning models. Accuracy follows the same final-answer scoring source as Table~\ref{tab:generalization_acc}, while CAR/PMF/HAR are computed with the 5-DAG best-alignment protocol. The same low-resource degradation pattern appears beyond Qwen3: low-resource traces generally have lower CAR/PMF and higher HAR than English or higher-resource target-language traces.

\section{English-Reasoning Baseline}
\label{app:english-reasoning-baseline}
We evaluate a practical answer-delivery baseline in which the model reasons in English inside \texttt{<think>} and then produces the user-facing explanation and final answer in the target language. Table~\ref{tab:english_reasoning_baseline} compares this setting with English reasoning/output and direct target-language reasoning on medium-difficulty PolyMath.

\begin{table}[H]
\centering
\scriptsize
\setlength{\tabcolsep}{3pt}
\renewcommand{\arraystretch}{1.08}
\begin{tabularx}{\columnwidth}{Xcc}
\toprule
Setting & sw Acc. (\%) & te Acc. (\%) \\
\midrule
English reasoning/output ($en\rightarrow en$) & 53.6 & 53.6 \\
Target-language reasoning ($en\rightarrow x$) & 4.8 & 8.0 \\
English reasoning + target-language output & 48.0 & 46.4 \\
\bottomrule
\end{tabularx}
\caption{English-reasoning + target-language-output baseline on medium-difficulty PolyMath with \texttt{Qwen3-4B}. Values are final-answer accuracy percentages.}
\label{tab:english_reasoning_baseline}
\end{table}

This baseline is strong for final-answer delivery, reaching 48.0\% and 46.4\% accuracy for Swahili and Telugu, close to English reasoning/output at 53.6\%. However, strict target-language compliance is only 29.6\% for Swahili and 21.6\% for Telugu, corresponding to 70.4\% and 78.4\% target-language drift. Thus, English reasoning followed by target-language output is useful in deployment, but it does not replace evaluating whether the reasoning trace itself remains reliable in the user's language.

\clearpage
\begin{table*}[t]
\noindent\begin{minipage}[t]{\columnwidth}
  \section{Correctness-Stratified DATG Results}
  \label{app:datg-correctness-full}
\end{minipage}%
\par\vspace{0.3em}
\centering
\tiny
\setlength{\tabcolsep}{2.0pt}
\renewcommand{\arraystretch}{0.70}
\resizebox{\textwidth}{!}{%
\begin{tabular}{llcccc cccc cccc}
\toprule
& & \multicolumn{4}{c}{\texttt{Qwen3-1.7B}} & \multicolumn{4}{c}{\texttt{Qwen3-4B}} & \multicolumn{4}{c}{\texttt{Qwen3-8B}} \\
\cmidrule(lr){3-6}\cmidrule(lr){7-10}\cmidrule(lr){11-14}
Pair & Diff. & Acc. (\%) & CAR & PMF & HAR & Acc. (\%) & CAR & PMF & HAR & Acc. (\%) & CAR & PMF & HAR \\
\midrule
\multirow{3}{*}{en-en} & High & 23.2 & .79/.37 & .64/.21 & .09/.16 & 41.6 & .86/.43 & .73/.27 & .04/.11 & 40.8 & .85/.40 & .72/.26 & .05/.09 \\
 & Med. & 36.8 & .88/.55 & .77/.37 & .04/.12 & 53.6 & .91/.64 & .83/.49 & .03/.04 & 53.6 & .94/.59 & .87/.45 & .01/.07 \\
 & Low & 81.6 & .98/.77 & .97/.59 & .00/.13 & 92.8 & .99/.80 & .98/.68 & .00/.07 & 94.4 & .98/.64 & .94/.37 & .01/.04 \\
\midrule
\multirow{3}{*}{en-fr} & High & 11.2 & .64/.29 & .45/.14 & .23/.24 & 31.2 & .81/.39 & .69/.25 & .08/.14 & 29.6 & .87/.33 & .77/.17 & .05/.20 \\
 & Med. & 31.2 & .75/.43 & .61/.27 & .06/.23 & 43.2 & .89/.62 & .79/.48 & .03/.10 & 50.4 & .90/.55 & .79/.41 & .01/.09 \\
 & Low & 73.6 & .94/.79 & .89/.66 & .02/.13 & 88.8 & .98/.77 & .97/.64 & .01/.10 & 92.0 & .99/.70 & .98/.53 & .01/.14 \\
\midrule
\multirow{3}{*}{en-zh} & High & 19.2 & .76/.35 & .59/.19 & .06/.20 & 36.0 & .85/.45 & .74/.29 & .02/.12 & 39.2 & .85/.34 & .72/.20 & .00/.14 \\
 & Med. & 36.8 & .86/.48 & .77/.31 & .04/.15 & 49.6 & .92/.67 & .82/.54 & .01/.06 & 54.4 & .93/.63 & .84/.47 & .01/.06 \\
 & Low & 77.6 & .96/.82 & .94/.70 & .02/.12 & 91.2 & .98/.86 & .95/.74 & .02/.02 & 92.0 & 1.00/.70 & 1.00/.43 & .00/.08 \\
\midrule
\multirow{3}{*}{en-ja} & High & 10.4 & .61/.27 & .44/.13 & .20/.32 & 23.2 & .86/.35 & .72/.20 & .06/.25 & 22.4 & .77/.29 & .64/.14 & .10/.20 \\
 & Med. & 18.4 & .81/.44 & .65/.27 & .09/.26 & 41.6 & .91/.60 & .80/.47 & .04/.12 & 39.2 & .84/.44 & .71/.25 & .03/.20 \\
 & Low & 67.2 & .96/.74 & .92/.57 & .03/.18 & 91.2 & .99/.81 & .99/.74 & .01/.08 & 91.2 & .99/.64 & .97/.46 & .01/.12 \\
\midrule
\multirow{3}{*}{en-ru} & High & 8.8 & .65/.34 & .49/.19 & .08/.26 & 32.8 & .77/.43 & .67/.26 & .04/.15 & 36.8 & .86/.37 & .72/.21 & .03/.15 \\
 & Med. & 32.0 & .81/.49 & .70/.33 & .07/.21 & 51.2 & .92/.63 & .82/.48 & .04/.09 & 49.6 & .89/.53 & .79/.35 & .01/.14 \\
 & Low & 75.2 & .94/.75 & .90/.54 & .03/.14 & 89.6 & .99/.85 & .98/.77 & .00/.07 & 94.4 & 1.00/.65 & .99/.36 & .00/.15 \\
\midrule
\multirow{3}{*}{en-ko} & High & 5.6 & .51/.25 & .33/.10 & .33/.31 & 24.0 & .80/.33 & .73/.18 & .10/.28 & 18.4 & .73/.25 & .59/.11 & .06/.23 \\
 & Med. & 17.6 & .75/.44 & .59/.26 & .11/.25 & 36.8 & .84/.61 & .72/.46 & .04/.14 & 36.0 & .81/.43 & .68/.25 & .04/.17 \\
 & Low & 74.4 & .94/.70 & .90/.47 & .04/.19 & 89.6 & .99/.83 & .97/.72 & .00/.11 & 87.2 & .98/.56 & .96/.27 & .00/.26 \\
\midrule
\multirow{3}{*}{en-id} & High & 9.6 & .71/.29 & .56/.15 & .11/.28 & 30.4 & .85/.38 & .77/.22 & .06/.17 & 28.0 & .87/.32 & .77/.17 & .03/.22 \\
 & Med. & 24.8 & .76/.45 & .62/.27 & .10/.20 & 47.2 & .92/.62 & .82/.48 & .03/.07 & 48.0 & .90/.50 & .81/.35 & .03/.20 \\
 & Low & 71.2 & .96/.71 & .94/.51 & .02/.16 & 92.0 & .99/.64 & .99/.42 & .01/.13 & 90.4 & .98/.63 & .96/.34 & .00/.21 \\
\midrule
\multirow{3}{*}{en-ms} & High & 5.6 & .76/.28 & .65/.14 & .12/.28 & 26.4 & .82/.39 & .69/.22 & .04/.19 & 27.2 & .80/.31 & .64/.17 & .03/.22 \\
 & Med. & 16.8 & .69/.45 & .56/.27 & .11/.18 & 44.8 & .90/.59 & .81/.45 & .02/.09 & 48.0 & .90/.47 & .81/.29 & .01/.15 \\
 & Low & 58.4 & .96/.78 & .92/.64 & .03/.10 & 92.0 & .99/.76 & .98/.57 & .01/.10 & 92.0 & .98/.49 & .97/.23 & .01/.25 \\
\midrule
\multirow{3}{*}{en-th} & High & 8.0 & .72/.25 & .57/.11 & .17/.31 & 16.8 & .74/.36 & .62/.20 & .08/.26 & 25.6 & .79/.29 & .68/.15 & .08/.28 \\
 & Med. & 22.4 & .75/.38 & .61/.20 & .11/.31 & 35.2 & .84/.58 & .74/.45 & .05/.17 & 48.8 & .88/.46 & .78/.31 & .01/.19 \\
 & Low & 72.0 & .96/.71 & .91/.49 & .02/.15 & 90.4 & .98/.81 & .96/.75 & .01/.05 & 92.0 & .99/.47 & .97/.18 & .00/.22 \\
\midrule
\multirow{3}{*}{en-bn} & High & 0.0 & --/.13 & --/.03 & --/.51 & 10.4 & .62/.20 & .50/.08 & .09/.36 & 12.8 & .78/.22 & .67/.08 & .09/.34 \\
 & Med. & 5.6 & .56/.24 & .34/.10 & .32/.46 & 16.0 & .79/.42 & .69/.27 & .13/.27 & 34.4 & .84/.36 & .74/.18 & .04/.27 \\
 & Low & 44.8 & .69/.55 & .56/.36 & .20/.30 & 84.0 & .95/.60 & .91/.44 & .04/.24 & 87.2 & .99/.57 & .99/.33 & .00/.33 \\
\midrule
\multirow{3}{*}{en-sw} & High & 3.2 & .18/.07 & .12/.03 & .75/.73 & 0.8 & .00/.06 & .00/.01 & 1.00/.68 & 2.4 & .64/.10 & .61/.03 & .00/.63 \\
 & Med. & 1.6 & .00/.09 & .00/.05 & .75/.77 & 0.0 & --/.12 & --/.04 & --/.69 & 16.0 & .82/.18 & .75/.08 & .13/.57 \\
 & Low & 27.2 & .65/.27 & .53/.17 & .26/.59 & 52.8 & .72/.44 & .65/.33 & .21/.48 & 59.2 & .95/.31 & .93/.16 & .06/.55 \\
\midrule
\multirow{3}{*}{en-te} & High & 0.8 & .00/.05 & .00/.01 & .00/.76 & 1.6 & .31/.10 & .25/.02 & .58/.56 & 4.0 & .66/.15 & .60/.04 & .23/.43 \\
 & Med. & 0.8 & .00/.08 & .00/.01 & 1.00/.70 & 4.0 & .35/.18 & .07/.05 & .27/.47 & 10.4 & .80/.27 & .71/.12 & .11/.38 \\
 & Low & 7.2 & .51/.26 & .33/.14 & .47/.61 & 72.0 & .81/.60 & .73/.49 & .14/.28 & 76.8 & .97/.49 & .94/.17 & .00/.31 \\
\bottomrule
\end{tabular}}
\caption{Correctness-stratified DATG results under $en\rightarrow x$ by language pair, difficulty, and model scale. Accuracy values are percentages. For CAR, PMF, and HAR, each cell reports C/I for correct and incorrect traces; ``--'' indicates no correct trace in that cell. DATG scoring does not use final-answer correctness.}
\label{tab:app_datg_correctness_full}
\end{table*}

\clearpage
\twocolumn[{%
  \begin{minipage}[t]{\columnwidth}
    \section{Test-Time Control Details}
    \label{app:controlled-execution}
  \end{minipage}%
  \par\vspace{0.4em}
  \begin{center}
    \scriptsize
    \setlength{\tabcolsep}{2.7pt}
    \renewcommand{\arraystretch}{1.05}
    \resizebox{\textwidth}{!}{%
    \begin{tabular}{lrrrrrrrrrrrrrrrrrr}
    \toprule
    \multirow{3}{*}{Method}
    & \multicolumn{9}{c}{Accuracy (\%) $\uparrow$}
    & \multicolumn{9}{c}{Decoded Cost} \\
    \cmidrule(lr){2-10}\cmidrule(lr){11-19}
    & \multicolumn{3}{c}{Low}
    & \multicolumn{3}{c}{Medium}
    & \multicolumn{3}{c}{High}
    & \multicolumn{3}{c}{Low}
    & \multicolumn{3}{c}{Medium}
    & \multicolumn{3}{c}{High} \\
    \cmidrule(lr){2-4}\cmidrule(lr){5-7}\cmidrule(lr){8-10}
    \cmidrule(lr){11-13}\cmidrule(lr){14-16}\cmidrule(lr){17-19}
    & sw & te & Avg
    & sw & te & Avg
    & sw & te & Avg
    & Tok. (M) & Retry\% & Trials
    & Tok. (M) & Retry\% & Trials
    & Tok. (M) & Retry\% & Trials \\
    \midrule
    Base
    & 45.6 & 64.8 & 55.2
    & 4.8 & 8.0 & 6.4
    & 4.8 & 3.2 & 4.0
    & 0.522 & 0.0 & 1.00
    & 1.772 & 0.0 & 1.00
    & 3.607 & 0.0 & 1.00 \\
    Loop-Retry
    & 74.4 & 74.4 & 74.4
    & 35.2 & 8.8 & 22.0
    & 25.6 & 6.4 & 16.0
    & 0.395 & 22.4 & 1.31
    & 1.809 & 50.4 & 1.90
    & 3.712 & 54.8 & 2.14 \\
    Formula-Retry
    & 87.2 & 75.2 & 81.2
    & 18.4 & 9.6 & 14.0
    & 8.0 & 4.0 & 6.0
    & 0.311 & 10.0 & 1.10
    & 1.937 & 46.8 & 2.00
    & 3.735 & 50.8 & 2.16 \\
    \bottomrule
    \end{tabular}}
    \vspace{-0.3em}
    \captionof{table}{All-difficulty test-time control results on Swahili and Telugu with \texttt{Qwen3-4B}. Token counts are decoded output tokens in millions, excluding prompt/input tokens.}
    \label{tab:control_all_difficulty}
    \vspace{0.8em}
\end{center}
}]

\paragraph{All-difficulty results and cost accounting.}
Test-time control cost reports decoded output tokens only. For retry-based
methods, decoded tokens include aborted continuations; for Formula-Retry, they
also include scaffold generation. Prompt/input tokens are not included, so this
is a decoded-token budget rather than total serving cost.
Table~\ref{tab:control_all_difficulty} extends the medium-difficulty comparison in Table~\ref{tab:control_main} by reporting Base, Loop-Retry, and Formula-Retry across all three difficulty levels. On low-difficulty problems, Formula-Retry is the strongest setting: it improves average accuracy from 55.2\% to 81.2\% and reduces decoded tokens from 0.522M to 0.311M. Loop-Retry also gives a large gain on this split, raising average accuracy to 74.4\%. On medium-difficulty problems, Loop-Retry gives the largest improvement, increasing average accuracy from 6.4\% to 22.0\% while keeping decoded cost close to Base, with 1.809M tokens compared with 1.772M. This gain mainly comes from Swahili, where accuracy rises from 4.8\% to 35.2\%. On high-difficulty problems, Loop-Retry remains the stronger control, improving average accuracy from 4.0\% to 16.0\%, again driven primarily by Swahili, where accuracy increases from 4.8\% to 25.6\%. By contrast, Formula-Retry gives only a small high-difficulty gain, from 4.0\% to 6.0\%. These results suggest that formula-level scaffolding is most effective on easier cases, whereas loop-targeted retry is more useful when low-resource traces degenerate on harder problems.

\subsection{Loop-Retry Trigger and Example}
\label{app:loop-retry}
\vspace{-0.5em}

Loop-Retry is an early-checkpoint repetition monitor, not a terminal failure
detector. It does not use the gold answer, answer correctness, Math-Verify,
DATG alignment, reference DAGs, language-compliance labels, selector labels, or
boxed-answer absence as retry triggers.

\paragraph{Early checkpoint.}
For a difficulty-specific generation budget \(B\), Loop-Retry inspects the
reasoning trace after the first quarter of the budget:
\[
C=B/4.
\]
Thus \(C=1024,2048,4096\) tokens for low, medium, and high difficulty,
respectively. In the reported high-difficulty run,
\(\texttt{check\_tokens}=4096\). We do not use a pre-check in the final setting
(\(\texttt{precheck\_tokens}=0\)).

\paragraph{Surface repetition detector.}
At the checkpoint, the detector computes repetition statistics over the most
recent 256 generated tokens. Let \(\mathrm{rep}_n\) be the repeated \(n\)-gram
rate:
\[
\mathrm{rep}_n =
\frac{\sum_g \max(0,\mathrm{count}(g)-1)}
     {N_n},
\]
where \(g\) ranges over token \(n\)-grams in the 256-token window and \(N_n\)
is the number of observed \(n\)-grams. We use \(\mathrm{rep}_{16}\),
\(\mathrm{rep}_{32}\), and the type-token ratio
\[
\mathrm{ttr}_{256}=\frac{|\mathrm{unique\ tokens}|}{256}.
\]
We also detect suffix motifs: a repeated suffix unit of length 4--32 tokens
repeated at least three consecutive times.

A partial trace is marked as a surface loop if
\[
\begin{aligned}
&\mathrm{rep}_{16}\ge 0.65
\quad\text{or}\quad
\mathrm{rep}_{32}\ge 0.30,\\
&\text{or a suffix motif is detected,}
\end{aligned}
\]
or if the softer joint condition holds:
\[
\mathrm{rep}_{16}\ge 0.50
\quad\text{and}\quad
\mathrm{ttr}_{256}\le 0.16.
\]

\paragraph{Mathematical progress guard.}
To avoid aborting legitimate long derivations, Loop-Retry also checks whether
the recent tail introduces new mathematical information. We inspect the last
1600 characters and extract equations, formula-like expressions, variable
assignments, and local number contexts. After normalization, if the tail
contains a mathematical span not previously seen in the prefix, we set
\(\texttt{math\_progress}=1\). The retry trigger fires only when a surface loop
is detected and \(\texttt{math\_progress}=0\). A partial trace that already
contains a boxed answer is not aborted. A partial trace shorter than \(0.8C\)
tokens is also not aborted, because it ended before the intended checkpoint.

\paragraph{Concrete Loop-Retry example.}
Loop-Retry keeps the original target-language prompt unchanged. If the first
continuation enters a repetitive loop at the checkpoint, that continuation is
discarded and the model resamples from the same prompt.

\begin{lstlisting}[style=promptstyle]
Problem:
Janet's ducks lay 16 eggs per day. She eats 3 for breakfast
and uses 4 for muffins. How many eggs are left after 7 days?

Target language: Swahili

Trial 1 continuation:
<think>
Kwanza tunahesabu mayai yanayobaki kila siku.
16 - 3 - 4 = 9.
Mayai yaliyobaki ni 9. Mayai yaliyobaki ni 9.
Mayai yaliyobaki ni 9...
[detector: repeated suffix, no new math progress]
=> abort and retry

Trial 2 continuation:
<think>
Kwanza tunahesabu mayai yanayobaki kila siku:
16 - 3 - 4 = 9.
Kwa siku 7:
9 \times 7 = 63.
</think>
\boxed{63}
\end{lstlisting}

\subsection{Formula-Retry Scaffold Prompt and Example}
\label{app:formula-prompt}
\vspace{-0.5em}

Formula-Retry uses a two-phase protocol. Phase 1 elicits a compact
English/symbolic pre-answer scaffold inside an execution-state block. Phase 2
continues the reasoning trace in the target language. The scaffold may contain
known quantities, variables, relations, and formula skeletons, but it must not
contain the final numeric result, a boxed answer, or a final-answer sentence.
The scaffold is generated once and kept fixed across retries; if the
continuation loops, only the target-language reasoning after
\texttt{</execution\_state>} is resampled.

\promptheader{Formula scaffold prompt}{lst:formula_prompt}
\begin{lstlisting}[style=promptstyle]
System message:
You are a mathematical reasoning assistant. Use a
two-phase reasoning protocol.
Phase 1 is a compact English/symbolic PRE-ANSWER
execution plan inside <execution_state>...</execution_state>.
It may use English words, formulas, variables, known
quantities, relations, and plan steps. The Plan section
should contain compact formula skeletons with unknown
results, e.g., "Calculate remaining eggs: 16 - 3 - 4 = ?".
You may include arithmetic operators, but all computed
intermediate and final numeric results must be masked as ?.
Do NOT compute the final numeric result in Phase 1. Do NOT
write a final answer, boxed answer, or any line named
Final/Answer. Do not use target-language prose in Phase 1.
Phase 2 is a step-by-step solution entirely in {language}.
Use the plan from Phase 1, perform the actual arithmetic in
{language}, and write both the reasoning trace and any
final-answer sentence in {language}. The only
English/symbolic part allowed in Phase 2 is the mathematical
expression inside \boxed{}. Do not write an English
explanation after </think>; after the reasoning trace,
output only \boxed{} or a short {language} final sentence
with \boxed{}. The execution state is a one-time planning
tool: never open or mention another execution-state block in
Phase 2.

User message:
{problem_statement}

Assistant prefill for Phase 1:
<think>
<execution_state>
Known:
-

After Phase 1 scaffold generation, the execution state is
closed and Phase 2 begins with the target-language starter:
</execution_state>
{target_language_starter}
\end{lstlisting}

\paragraph{Concrete Formula-Retry example.}
The scaffold contains only answer-free structure. In a retry, the same scaffold
is reused and only the continuation after \texttt{</execution\_state>} changes.

\begin{lstlisting}[style=promptstyle]
Problem:
Janet's ducks lay 16 eggs per day. She eats 3 for breakfast
and uses 4 for muffins. How many eggs are left after 7 days?

Phase 1 scaffold:
<think>
<execution_state>
Known:
- eggs_per_day = 16
- eaten_per_day = 3
- muffins_per_day = 4
- days = 7

Plan:
- remaining_per_day = 16 - 3 - 4 = ?
- total_remaining = remaining_per_day * 7 = ?
</execution_state>

Trial 1 after scaffold:
Kwanza tunatumia mpango uliopo.
16 - 3 - 4 = 9.
9 ni mayai yaliyobaki. 9 ni mayai yaliyobaki...
[detector: loop, no new math progress]
=> keep scaffold, retry only after </execution_state>

Trial 2 after same scaffold:
Kwanza tunahesabu mayai yanayobaki kila siku:
16 - 3 - 4 = 9.
Kisha kwa siku 7:
9 \times 7 = 63.
</think>
\boxed{63}
\end{lstlisting}

\subsection{Shared Retry Policy and Decoding}
\label{app:retry-decoding}
\vspace{-0.5em}

\paragraph{Retry policy.}
Each retry method uses at most five trials. If the checkpoint trigger fires
before the final trial, the current partial continuation is discarded and the
model is resampled from the same prompt context. For Formula-Retry, the
\texttt{<execution\_state>} scaffold is generated once and kept fixed; only the
reasoning continuation after \texttt{</execution\_state>} is resampled. If the
trigger still fires on the final trial, the final trial is accepted and marked
as an unhealthy forced accept; no additional retry is performed.

All reasoning continuations use temperature \(0.6\), top-\(p=0.95\), and the
maximum budget \(B\). The first trial uses repetition penalty \(1.08\), and
retry trials use repetition penalty \(1.13\). Formula-scaffold generation uses
temperature \(0.2\), top-\(p=0.95\), repetition penalty \(1.02\), and a maximum
of 4096 tokens, stopping at \texttt{</execution\_state>}. We use vLLM with
\(\texttt{max\_num\_seqs}=16\), \(\texttt{max\_num\_batched\_tokens}=8192\),
and GPU memory utilization \(0.82\). All test-time control experiments are run with three independent sampling seeds, 42, 215, 316.

\paragraph{Retry prompts.}
Base and Loop-Retry use the same target-language solver prompt and the same
assistant-side direct-first starter; Loop-Retry changes only the test-time
sampling procedure when the repetition trigger fires. Formula-Retry uses the
two-phase scaffold prompt in
Appendix~\ref{app:formula-prompt}. Formula-Retry keeps the sanitized
\texttt{<execution\_state>} fixed across retries and resamples only the
continuation after \texttt{</execution\_state>}.

\section{Environment and Hyperparameters}

All experiments are conducted on a server equipped with 8 NVIDIA RTX A6000 GPUs. 
Model inference is performed with \texttt{vLLM} \citep{Kwon2023vllm}.

\end{document}